\RequirePackage{fix-cm}
\documentclass[twocolumn, captions=
figuresignature]{svjour3}          
\smartqed  

\usepackage{natbib}
\usepackage[normalem]{ulem}
\usepackage{mathtools}
\usepackage{algorithm}
\usepackage{graphicx}
\usepackage[caption=false]{subfig}
\usepackage[noend]{algpseudocode}
\usepackage{multirow}
\usepackage{hhline}

\usepackage{newtxtext,newtxmath}

\newcommand{\R}{\mathbb{R}}
\usepackage{varwidth}

\usepackage{pdfrender}

\usepackage{bm}
\newcommand*{\boldgreek}[1]{%
  \textpdfrender{%
    TextRenderingMode=FillStroke,%
    LineWidth=.35pt,%
  }{#1}%
}
\newcommand\norm[1]{\left\lVert#1\right\rVert}
\algtext*{EndWhile}
\algtext*{EndIf}
\DeclareMathOperator*{\argmax}{argmax} 

\usepackage{xcolor}

\usepackage{etoolbox}
\newcommand*{\affaddr}[1]{#1} 
\newcommand*{\affmark}[1][*]{\textsuperscript{#1}}

\usepackage{color,soul}

\begin{document}

\title{Heterogeneous robot teams for modeling and prediction of multiscale environmental processes 
}


\author{Tahiya Salam   \protect\affmark[1]      \and
        M. Ani Hsieh \protect\affmark[1] 
}


\institute{Tahiya Salam \at
              \email{tsalam@seas.upenn.edu}           
           \and
           M. Ani Hsieh \at
              \email{mya@seas.upenn.edu}           \\
             \and
             \affaddr{\affmark[1]GRASP Laboratory \\ University of Pennsylvania, Philadelphia, USA}
}

\date{Received: date / Accepted: date}

\maketitle

\begin{abstract} 
This paper presents a framework to enable a team of heterogeneous mobile robots to model and sense a multiscale system. We propose a coupled strategy, where robots of one type collect high-fidelity measurements at a slow time scale and robots of another type collect low-fidelity measurements at a fast time scale, for the purpose of fusing measurements together. The multiscale measurements are fused to create a model of a complex, nonlinear spatiotemporal process. The model helps determine optimal sensing locations and predict the evolution of the process. Key contributions are: i) consolidation of multiple types of data into one cohesive model, ii) fast determination of optimal sensing locations for mobile robots, and iii) adaptation of models online for various monitoring scenarios. We illustrate the proposed framework by modeling and predicting the evolution of an artificial plasma cloud. We test our approach using physical marine robots adaptively sampling a process in a water tank.

\end{abstract}

\section{Introduction}
\label{sec:intro}
Multi-robot systems have a long-standing history of being used to solve problems that are distributed in time and space, with increased efficiency \citep{Parker2003TheRobots}. In the past, contributions to multi-robots teams have come from swarm intelligence, optimal control and optimization, motion planning and coordination, task allocation, distributed perception and estimation, decision making, and learning. While many existing works consider teams composed of identical robots, i.e., homogeneous teams, there is an increased interest in the use of heterogeneous teams in recent years \citep{Prorok2015FastRobots, Prorok2016FormalizingRobots, Vasilijevic2015HeterogeneousSurvey, Manjanna2018HeterogeneousSampling, Manderson2019HeterogeneousSampling, Maini2018Visibility-BasedTeam}. Heterogeneity in multi-robot teams refers to the differing physical or behavioral characteristics amongst agents. The impact of heterogeneity has been studied in the context of cooperative sensing and coordination of multi-agent robotic systems. Complex tasks such as search and rescue often benefit or may require robots with different capabilities to coordinate and collaborate to accomplish the task. Advantages of heterogeneous robot teams over their homogeneous counterparts include the ability to introduce varying degrees of cooperation amongst team members, sharing information from different sources, faster exploration times, and increased robustness in the event of failure. These advantages are most evident when deploying robot teams to track dynamic and multiscale phenomena that varies in both space and time, {\it e.g.}, oil spills, forest fires, and air pollution. For example, more agile aerial robots coupled with ground vehicles that can get closer to a target of interest may provide better situational awareness in complex environments but still maintain widespread surveillance of the area. While such heterogeneous teams clearly bring significant advantages over homogeneous ones, heterogeneity brings new challenges in coordination, decision-making, planning, network connectivity, and information exchange. Furthermore, as robots permeate all aspects of our lives, {\it e.g.}, self-driving cars, package delivery drones, autonomous environmental monitoring systems for wildlife preservation, to name a few, multi-robot interactions will become more ubiquitous. As such, the question of how best to coordinate these agents and better leverage heterogeneity and diversity in capabilities for complex tasks must be addressed. 

In this work, we focus on the task of optimal sensing and online map merging for a team of heterogeneous robots tracking a multiscale dynamic process. In many environmental monitoring applications, the objective is to enable the robot team to map and track dynamic processes such as fluid flows, animal and insect swarms, forest fires, dispersion of airborne pollutants, and crowds in complex environments \citep{Dunbabin2012RobotsApplications, Singh2010ModelingAmarjeet, Manderson2019HeterogeneousSampling, Strogatz2001NonlinearNonlinearity}.  These processes are often multiresolution in the sense that they exhibit complicated patterns across various space-time scales. The many disparate space-time scales makes it difficult to estimate the process with a single, or even single type of, robot. A robot team with heterogeneous mobility and sensing capabilities can be more adept at collecting and fusing the multiresolution data needed to model and track these multiscale dynamic processes. The robots can represent these processes as maps and use the maps to adapt their sensing locations to improve them in an online fashion. In the online model adaptation component, the previously derived models of the dynamic process are updated with the acquisition of new data. In multi-robot modeling, the environment is often assumed to be fixed. In these static environments, the time scales tend be very long and changes in the environment are mostly discrete, {\it e.g.} an office door is either open or closed. However, this is not the case in modeling multiresolution dynamic processes, as they are both varying in time and also exhibiting different time scales. In these dynamic environments, the model must account for the added complexity of parts of the model changing. Thus, an online model adaptation is essential for quickly incorporating new process data to maintain model quality. 

\begin{figure*}\centering
\includegraphics[width=0.9\textwidth]{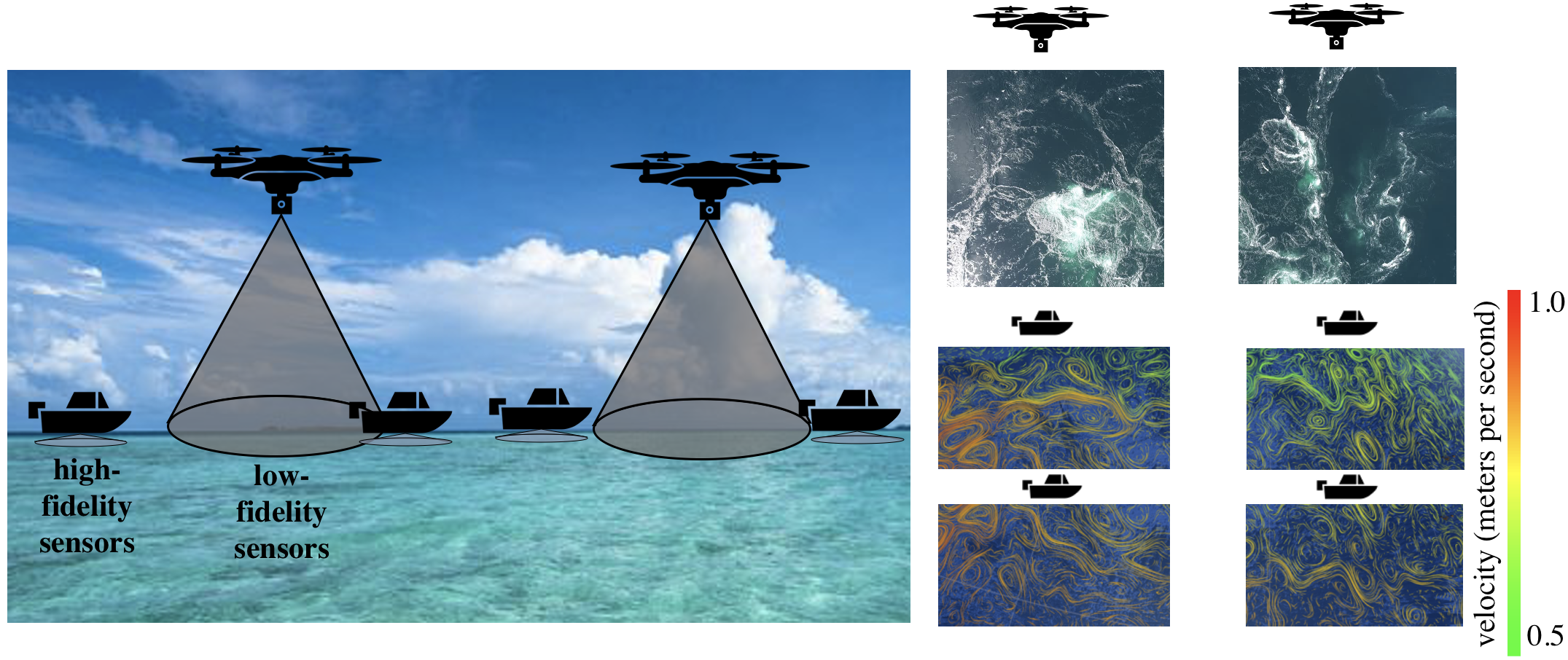}
\caption{Heterogeneous robots collecting different sensing information work to create a cohesive model of a time varying environment. Aerial vehicles collect low-fidelity sensor measurements, such as overhead images, over a wide area, and marine vehicles collect high-fidelity sensor measurements, such as current speeds, over a small area. Sensor measurements are unified into one model for estimation and prediction of a time varying process}
\label{fig:overview}       
\end{figure*}

The core contribution of this work is addressing how to cohesively model an unknown, complex, spatiotemporal environment using a team of heterogeneous robots. We define this broad class of environments simply as dynamical systems to allow us to use tools from dynamical systems theory to model the environment. Such a comprehensive problem statement encompasses many scenarios of interest for robotic applications. For example, this can include incorporating various types of robots with diverse sensing and mobility modalities to build a unified map of the environment, such as ground and aerial vehicles jointly mapping temperature profiles or marine and aerial vehicles jointly mapping ocean currents, as in Fig. \ref{fig:overview}. To that end, these problems require tight coordination amongst robots to ensure the model is accurate, consistent, and updated online. 

At the same time, this work leverages the unique strengths of heterogeneous multi-robot teams without explicitly formulating a task allocation problem that would be computationally intractable \citep{Korsah2013AAllocation}. This alleviates the need to explicitly enumerate the requirements of the task and match them to the traits of the robots. In the proposed strategy, the notion of heterogeneity is encapsulated through fusion of multiscale data from disparate sources into a cohesive, unified model. This work extends the techniques for adaptive modeling and prediction through modal analysis techniques to heterogeneous multi-robot systems. 

The proposed online mapping and task allocation strategy is rigorously formulated, with computational complexity analysis provided. The strategy is also accompanied by simulation studies using data of a multiresolution dynamic processes. We present the development of the online task allocation and data assimilation strategy. Experimental studies using mixed-reality teams, consisting of simulated aerial vehicles and actual marine vehicles were conducted using simulated data for a barium plasma cloud. This enables us to successfully address and present experimental scenarios with truly multiscale data of processes that exhibit complex, unpredictable behaviors.

To summarize, the technical novelty of this work lies in the introduction of a novel, heterogeneous, multi-robot online modeling and estimation framework, which determines optimal sensing locations from the estimated learned model of a multiscale process representing the environment, in order to improve the model accuracy. The remainder of the paper is organized as follows: Sect. \ref{sec:related-works} overviews the related works in heterogeneous multi-robot systems. Sect. \ref{sec:problem-statement} outlines the problem definition, the assumptions, and the mathematical foundations for the algorithms used. The proposed approach and main contributions are presented in Sect. \ref{sec:proposed-approach}. Computational analysis and simulation-based evaluations are presented in Sect. \ref{sec:simulation-analysis}. Experimental evaluation studies are demonstrated in Sect. \ref{sec:experimental-validation}. Sect. \ref{sec:conclusion-future} ends with conclusions and future research directions.



\section{Related works}
\label{sec:related-works}
In this section, we review the state-of-the-art in heterogeneous multi-robot systems, with special attention to current solutions for multi-robot modeling and data fusion with heterogeneous data.
\subsection{Heterogeneous multi-robot systems}
In the multi-robot literature, heterogeneity is often framed as a task allocation problem \citep{Jones2006DynamicallyTasks, Rossi2009SimultaneousRobots, Korsah2013AAllocation, Prorok2015FastRobots, Prorok2016FormalizingRobots, Bae2019ACost, Liang2019Multi-targetSpace, Notomista2019AnSystems}.  Starting with the classification of multi-robot coordination by \cite{Gerkey2004ASystems}, followed by its  categorization by \cite{Korsah2013AAllocation}, and reviewed by \cite{Khamis2015Multi-robotState-of-the-art}, existing work has mainly focused on coordination in the form of task-to-trait allocation. In this framework, robots possess traits that allow them to complete task specifications. These traits can include mobility capabilities, on-board sensors, and/or other characteristics that allow the robots to make progress on or complete specified tasks. Though the premise of traits has been used in prior works, the terms were first formalized by \cite{Prorok2015FastRobots} for heterogeneous teams. According to \cite{Prorok2015FastRobots}, robots possess a set of traits that allow them to accomplish tasks where these tasks can also be decomposed in terms of traits. Such a discretization of both the robots' capabilities and the tasks' requirements may not always be feasible, especially in situations where the tasks cannot be easily decomposed and/or robots capabilities may not exactly fulfill the tasks' requirements. Even in situations where one considers the distributions of traits rather than their binary existence, such a formulation may ignore some of the original requirements of the task \citep{Prorok2015FastRobots, Prorok2016FormalizingRobots}. For example, in this work the task is to reconstruct and predict the evolution of the environment using heterogeneous robots.  The discretization of this task cannot be easily decomposed into the sensing capabilities of the individual robots, as all of the collective information must be leveraged to accomplish the task. 

\subsection{Multi-robot sensing and modeling}
The question of modeling with multi-robot teams has been a topic of interest in the robotics community for many years \citep{nashashibi1992indoor}. Heterogeneity in sensing and modeling has been explored in the coverage control literature. Coverage control encompasses a type of optimal sensor allocation problem. Coverage control has a long-standing history of using multi-robot teams to sense environmental features through a density function \citep{Cortes2004CoverageNetworks}. The density function used in coverage control techniques can be thought of as a model of the process of interest. For heterogeneous sensing modalities, each sensing modality is represented as a different density function and the optimal sensor allocation is solved as an optimization problem over these combined density functions \citep{Santos2018CoverageCapabilities, Santos2018CoverageCommunications, Sadeghi2019CoverageRobots}. In the cases where the model is unknown {\it a priori}, robots perform a distributed estimation of unknown density functions \citep{Julian2012DistributedApproach}. However, these techniques may not be suited for fast or unknown dynamics where the density function changes rapidly in time or is unknown to the robots. Lastly, these techniques neglect how to assimilate new sensor measurements from robots. 

Gaussian Process (GP) modeling has been used in multi-robot systems for modeling, determining optimal sensor placement, prediction, and path planning \citep{Krause2007NonmyopicApproach, Krause2008Near-optimalStudies, Singh2009EfficientRobots, Singh2010ModelingAmarjeet, Xu2011MobileObservations}. GPs are a non-parameteric model that can be derived using data \citep{Rasmussen2004GaussianLearning}, which makes them extremely appealing for modeling the environment using data collected by robots. These techniques have been extended and applied to heterogeneous teams, where heterogeneity refers to robots of different types ({\it e.g.} aerial and ground) or different sensing modalities ({\it e.g.} water quality and sampling apparatus) \citep{Manderson2019HeterogeneousSampling, Manjanna2018HeterogeneousSampling}. In these scenarios, robots are delegated specialized roles that require fulfilling disparate tasks in parallel, such as a robot building a GP model while the other samples. Requiring one of the robots to build the GP model means the robot runs into the same problems associated with standard robotic GP modeling applications. For one, determining the appropriate kernel function that actually captures the complex spatiotemporal relationships in the environment is a challenging task and still remains an open question \citep{Singh2010ModelingAmarjeet}.  Additionally, the size of the covariance matrix increases as more measurements are collected making its inverse expensive to compute. Some studies propose efficient methods for using only a subset of observations instead of all observations to compute the covariance matrix \citep{Xu2011MobileObservations}. However, these methods impose restrictions on the covariance functions. 

Modal analysis techniques have been used to study spatiotemporal systems and extract the important physical features \citep{Taira2017ModalOverview}. Modal analysis techniques for fluid flows, finance, video processing, and robotics have been well studied \citep{Mezic2005, Rowley2009SpectralFlows, Schmid2010DynamicData, Tu2013OnApplications, Jovanovic2014Sparsity-promotingDecomposition, Brunton2016ExtractingDecomposition, Zhang2019OnlineSystems, Matsumoto2017On-the-flySquares, Manohar2018Data-DrivenPatterns, Manohar2019OptimizedDynamics, Salam2019}. Dynamic mode decomposition (DMD) is a modal analysis technique that decomposes time series data into modes, where each mode has a characteristic frequency and growth or decay rate \citep{Schmid2010DynamicData}. DMD relies only on time series data and does not require governing equations. This method combines advantages from two powerful data analytic tools: power spectral analysis for temporal analysis and principal components analysis (PCA) for spatial analysis. Additionally, DMD can be used for prediction of future states, optimal sensor placement, estimation, and control, which can be useful in many robotic applications \citep{Taira2017ModalOverview, Manohar2018Data-DrivenPatterns}. Despite DMD and similar techniques being used for robotic applications, most uses of DMD in robotics have focused on control and not modeling \citep{Berger2015EstimationDecomposition, Folkestad2020ExtendedControl}.

\subsection{Data fusion for heterogeneous multi-robot teams}
Map merging techniques are methods for fusing information into a global map, when multiple robots have been used to explore the environment and collect information. Map merging focuses mainly on map matching, the problem of finding the correspondences between maps, and map fusion, the problem of merging the data from maps in the case where there is a known or estimated correspondence between the maps. For a review of the state of art in map merging techniques, we refer the interested reader to the work by \cite{Andersone2019HeterogeneousArt}. Most map merging techniques focus on homogeneous robots. Works focusing on heterogeneous robots have mostly focused on map fusion for occupancy grid maps of various scales \citep{Topal2010ASystems, Park2016MapFeatures, Ma2016MergingRegistration, Shahbandi20192DDecomposition}. However, there is no universal solution to map merging, especially for heterogeneous robots, due to the varying degrees of heterogeneity in format, sensing type, and scale of data. As such, there are still many open challenges to merging of data and developing coordination strategies for optimal sensing and modeling by heterogeneous robot teams. 

\subsection{Contributions} The work presented in this paper avoids the formulation of heterogeneity as a task allocation problem and instead considers multiple tasks that must be solved sequentially using all available traits of the team. Unlike coverage control and GP techniques, the proposed approach explicitly considers the impact of heterogeneous data sources and online data assimilation in building a single map. Likewise, map merging is still not well studied for robotic frameworks leveraging heterogeneity. As mentioned previously, the core contribution focuses on modeling and optimal sensing tasks and requires the coupling of different resources from the heterogeneous robots, while addressing the challenges of map merging and online updates.

\section{Problem statement}
\label{sec:problem-statement}
The main contribution of this work is the synthesis of techniques from nonlinear dynamical systems theory and robotics in the development of a distributed estimation and sensing framework for environmental monitoring by heterogeneous robot teams. The proposed framework enables us to model the environment in principled ways while accounting for the constraints and challenges posed by multi-robot systems. In the proposed framework, heterogeneous robot capabilities are abstracted through the reduced-order models created by the robots. For example, aerial vehicles generally have wider fields of view, while marine vehicles are able to directly sample the process. Thus, the heterogeneity of the robots is reflected through their collection of multiresolution data. Aerial vehicles collect data with low spatial resolution at higher frequency and marine vehicles collect data with high spatial resolution at slower frequency. 

\paragraph{Problem statement:} This work extends the techniques for adaptive modeling and prediction through dimensionality reduction and learning techniques to heterogeneous multi-robot systems. Accomplishing this requires combining data from different sources and of varying resolutions into one cohesive model, while maintaining the independence of data sources. In order to resolve the various temporal rate and spatial resolutions, the proposed scheme ensures that lower resolution data collected at faster rates are interpolated to the same resolution as the higher resolution data collected at lower rates. The multi-resolution data is aggregated in a centralized fashion. Model-learning techniques are then employed on this resolved data to obtain a map capable of capturing the spatiotemporal dynamics of the system, which is then used to guide the allocation of the aerial and marine robots. The overall problem is comprised of three main aspects: 1) model estimation and uncertainty quantification, 2) uncertainty-reducing task assignment, and 3) online adaptation of models, described in Sections \ref{subsec:model-est}, \ref{subsec:task-assgn}, and \ref{subsec:online-adpt} respectively.

\subsection{Background and assumptions}\label{subsec:background}
The proposed framework employs DMD for model estimation and uncertainty quantification. For a more detailed exposition on DMD and its relationship to dynamical systems theory for nonlinear differential equations, we refer the interested reader to \citep{Mezic2005, Rowley2009SpectralFlows, Budisic2012AppliedKoopmanism}.

Given $T + 1$ sequential snapshots $\{ \mathbf{x}(0), \mathbf{x}(1), \dots, \mathbf{x}(T)\}$, let each $\mathbf{x}(t) \in \R^N$ represents field values at $N$ spatial locations at time $t$. Then, assume there is a linear dynamical relationship
\begin{equation} \label{eq:x1_A_x0}
    \mathbf{x}(t+1) = \mathbf{Ax}(t)
\end{equation}
between two consecutive snapshots. The matrix $\mathbf{A}$ is typically extremely high-dimensional. DMD allows us to learn a reduced-order model of the matrix $\mathbf{A}$ that captures the important spatial and temporal characteristics of the data. Let
\begin{gather}
    \mathbf{X} = [\mathbf{x}(0) \quad \mathbf{x}(1) \quad \dots \quad \mathbf{x}(T-1) ] \\ 
    \mathbf{Y} = [ \mathbf{x}(1) \quad \mathbf{x}(2) \quad \dots \quad \mathbf{x}(T)]
\end{gather}
be defined as the snapshot matrices, a matrix with its columns as snapshots of collected data.
Then, in matrix notation, Eq. \eqref{eq:x1_A_x0} can then be written as $\mathbf{Y} = \mathbf{A} \mathbf{X}$. Thus, DMD is an approximate eigendecomposition of the operator 
\begin{equation}\label{eq:A-matrix}
    \mathbf{A} = \mathbf{Y} \mathbf{X} ^\dagger,
\end{equation}
where $^\dagger$ is the pseudoinverse operator. 

While this is a linear approximation, many of the papers cited above mention the idea that DMD is able to characterize nonlinear dynamics through an analysis of some approximating linear system \citep{Tu2013OnApplications}. \cite{Rowley2009SpectralFlows} established the connection between DMD and nonlinear dynamical systems and helped explain the validity of DMD when applied to nonlinear systems. 

Given that the DMD operator $\mathbf{A}$ may be high dimensional and difficult to compute, we use an efficient algorithm that computes the eigendecomposition of $\mathbf{A}$ through a low-dimensional approximation \citep{Schmid2010DynamicData}. For this procedure, we compute a singular value decomposition (SVD) on $\mathbf{X}$ such that
\begin{equation} \label{eq:x-decomp}
     \mathbf{X} = \mathbf{U} \bm{\Sigma} \mathbf{W}^T
\end{equation}
where $\bm{\Sigma}$ is an $r \times r$ diagonal matrix of non-zero singular values $\{\sigma_0, \sigma_1, \dots , \sigma_{r-1} \}$ and $r$ is the rank of the snapshot matrix. Then, combining Eqs. \eqref{eq:A-matrix} and \eqref{eq:x-decomp} gives $\mathbf{Y} = \mathbf{A} \mathbf{U} \bm{\Sigma} \mathbf{W}^T$ and therefore an approximation of $\mathbf{A}$ can be given by 
\begin{gather}
    \hat{\mathbf{A}} = \mathbf{U}^T \mathbf{A} \mathbf{U} \\
    \hat{\mathbf{A}} = \mathbf{U}^T \mathbf{Y} \left( \mathbf{X} \right) ^\dagger \mathbf{U} = \mathbf{U}^T \mathbf{Y} \mathbf{W} \bm{\Sigma}^{-1} \mathbf{U} \label{eq:projected-A}.
\end{gather}
Given the operator $\hat{\mathbf{A}}$, we can compute its eigendecomposition as
\begin{equation} \label{eq:A-decomp}
    \hat{\mathbf{A}} \mathbf{V} = \mathbf{V} \mathbf{\Lambda}
\end{equation}
where $\mathbf{V}$ contains the eigenvectors of $\hat{\mathbf{A}}$ and $\mathbf{\Lambda}$ is a diagonal matrix of the eigenvalues, $\lambda_i$, of $\hat{\mathbf{A}}$.

The DMD modes can then be computed as 
\begin{equation} \label{eq:dmd-modes}
    \mathbf{\Phi} = \mathbf{Y} \mathbf{W} \bm{\Sigma}^{-1} \mathbf{V}
\end{equation}
where each column, $\boldgreek{\phi}_i$ of $\mathbf{\Phi}$ is the DMD mode corresponding to the eigenvalue, $\lambda_i$ found in Eq. \eqref{eq:A-decomp}. The reconstruction of the data is then written as
\begin{equation}\label{eq:dmd-reconstructed}
    \hat{\mathbf{x}}(t) = \mathbf{\Phi}\mathbf{\Lambda}^t \bm{\alpha}
\end{equation}
where $\bm{\alpha}$ is computed using least-squares for $\mathbf{x}(0) = \mathbf{\Phi}\bm{\alpha}$. Thus, the spatial modes are captured by the vectors $\boldgreek{\phi}_i$ and the temporal dynamics of the spatial modes are captured by $\lambda_i$. The magnitude of the vector $\boldgreek{\phi}_i$ represents the spatial correlations between the set of locations, and the magnitude and phase components of the eigenvalue $\lambda_i$ represents the rate of growth/decay and frequency of oscillation of the corresponding spatial mode $\boldgreek{\phi}_i$.

Given a workspace, $\mathcal{W} \subset \mathbb{R}^d$ for $d \in \{2, 3\}$ discretized into $N$ points, each point can be denoted as $p_i \in \mathbb{R}^d$, for $i \in  I = \{1, \dots, N\}$, where $I$ is the index set of the discretization. That is, each point $p_i$ can be uniquely identified by an index $i$. The time varying process of interest $\mathcal{P}$ is observed over the workspace $\mathcal{W}$ and sensed over the times $\{0, \dots, \mathcal{T} \}$. Each point $p_i$ has an associated weight, $w_{i}$, corresponding to the informativeness of that location with respect to mapping $\mathcal{P}$ and some specified statistical criterion. Robots are able to collect various sensor measurements over a set of points $S_i$, where $S_i \subseteq {I}$, depending on their sensing quality, sensing radius, and location $p_i$ in the workspace. Let $m$ be the number of robots on a team with one type of sensing modality, $m'$ be the number of robots with a different type of sensing modality, and the total number of robots be $M = m + m'$. Each robot $j$, for $j \in \{1, \dots, M \}$, collects information at different time scales $T_{j}$, where $T_{j} \subseteq \{0, \dots, \mathcal{T}\}$, depending on the robot's sensing capabilities. This means that each robot $j$ at location $p_j$ has its unique sensing locations $S_j$ and collects data at the times $T_j$. In this case, the sensing locations are non-overlapping by the allocation of the robots but the data collection times may be overlapping. 

\subsection{Model estimation and error quantification}\label{subsec:model-est}
In order for the heterogeneous robots to build a model of the previously unknown area, the proposed strategy needs to provide a framework for a compact, meaningful representation of the environment. Hence, the heterogeneous data collected from the robots should be unified to reconcile the differences in spatial and temporal scales, build a cohesive model, and extract the key features of the environment from the model. The multiscale, multiresolution data collected should be combined as input to the algorithm that constructs a model. 

The environment is discretized into points within the workspace and represented as a state $\mathbf{x}$, where the state changes in time. This representation of the environment as a states lends to high-dimensional systems. Given that the high-dimensionality of these systems is computationally prohibitive, the model derived by the robots should have lower dimensionality and elucidate the key spatial and temporal characteristics of the process to allow for planning. Selecting a model that highlights the meaningful features of the various spatial and temporal properties of the dynamical system, such as DMD, allows for not only model reconstruction but also model-informed planning of optimal sensing locations. In order to so, the selected methodology for modeling has to support a measure of informativeness with respect to sensing information, wherein collecting data in different locations has measurably different effects on model quality. Additionally, the model should reduce the error in reconstruction from limited sensing data collected from mobile robots.

\subsection{Uncertainty-reducing task assignment}\label{subsec:task-assgn}
The optimal sensor allocation problem is formulated as an optimization over the developed map, specifically the spatial and temporal components captured in $\mathbf{\Phi}$ and $\mathbf{\Lambda}$, as seen in Eq. \eqref{eq:projected-A} and \eqref{eq:dmd-modes}. During this phase, the robots are assigned to the optimal sensor locations that lowers the map uncertainty. This couples the coordination of the robot team, a task allocation problem, to the quality of completing the task, increasing model quality.

Planning where mobile robots should go next to improve the model is essential in the online planning framework. Given that the robots are operating in an extremely high dimensional space and only have limited sensing radius and capacity, this entails finding a small subset of optimal sensing locations such that the uncertainty is reduced or the information gain is maximized. Given informativeness of optimal sensing locations, determination of where robots should go to collect the sensing measurements is NP-hard. However, instead of task to trait assignment, as in the heterogeneity literature \citep{Korsah2013AAllocation}, this task allocation problem is restrained to the domain of assigning sensing regions to robots, as in the spirit of coverage problems.

To begin, the model is unknown and the robots collect sensor measurements to build an initial model of the environment. After an initial model is built, the aim is to determine which locations of radius $k$ are expected to lead to the most uncertainty reduction in estimating the model. These locations are reevaluated and redetermined after more sensing data is collected and the models are updated.

For every point $p_i \coloneqq (x,y) \in \mathcal{W}$, we can define a sensing region, $S_i$, from that location $p_i$ as a collection of points, $\{ p_j \coloneqq (x', y') \}$, such that
\begin{equation} \label{eq:sensing-region}
    S_i \coloneqq \{p_j \enskip | \enskip  ||(x - x', y - y')^{\top}|| \leq k \}.
\end{equation}

The cumulative weight $W_{i}$ of each sensing regions $p_i$ is then $W_{i} = \sum\limits_{p_j \in S_i} w_{j}$.

\begin{figure*}[ht!]
\centering
\subfloat[ ]{
  \includegraphics[width=53mm]{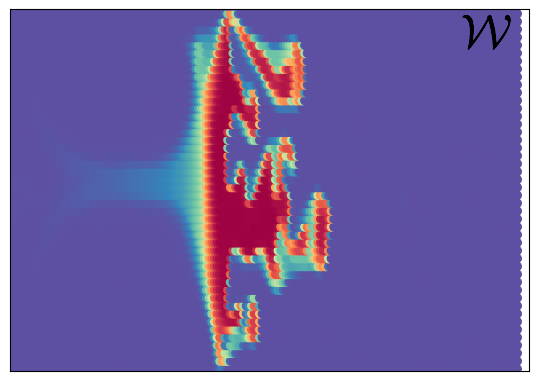}
}
\subfloat[ ]{
  \includegraphics[width=53mm]{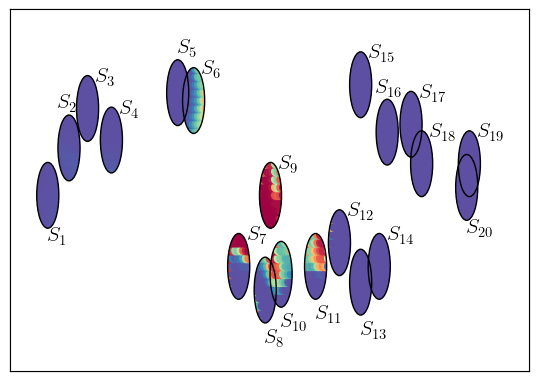}
}
\subfloat[ ]{
  \includegraphics[width=53mm]{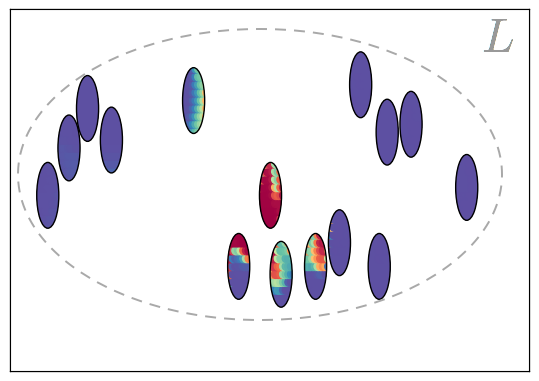}
}
    \caption{Depiction of selecting $15$ candidate sensing regions from workspace. a) The workspace $\mathcal{W}$ is the space over which the time varying process $\mathcal{P}$ is observed. b) The process $\mathcal{P}$ can be divided into candidate sensing regions, $S_i$, where for each set represents a set of points that a robot would be able to sense if they were located in the center of the region. c) In order to construct a valid set $L$ that contains all of the points from one possible assignment of robots to regions, overlapping sets $S_5$, $S_8$, $S_{13}$, $S_{18}$, and $S_{19}$ have been removed. Note, that there are many possible sets $L$, such as the set of points resulting from removing $S_{10}$ instead of $S_8$. The whitespace $\mathcal{W} \setminus L$ is inferred using a model and the measurements in the set $L$} \label{fig:sensing-sets}
\end{figure*}

Define $L$ as a set containing the elements within the sets $S_1, S_2, \dots, S_m$ that are arbitrary sensing regions, such that for any two sets $S_i$ and $S_j$ with elements in $L$, $S_i \cap S_j = \varnothing$. Thus, $L$ corresponds to all of the locations in $m$ non-overlapping sensing regions. Note, there are multiple ways to satisfy the definition of $L$. Intuitively, this means $L$ contains all of the points for which sensing data is being collected over a specific allocation of the robots to certain locations. This is shown in detail in Fig. \ref{fig:sensing-sets}. 

The values of the process $\mathcal{P}$ can be inferred at all locations without available sensor measurement, defined as the points in the set $\mathcal{W} \, \setminus \, L$, for a specific $L$ given the choice of model and an appropriate estimation scheme.

The optimization problem is then 
\begin{equation} \label{eq:opt}
    L^{*} \coloneqq \argmax\limits_{L} \sum\limits_{S_j \subset L} W_{j},
\end{equation} 
as in finding the set of optimal locations $L^{*}$ such that the cumulative weight over these $m$ sensing locations is maximized.

This is a combinatorial optimization problem for $m$ instances over the entire workspace of dimension $N$, where $N$ is a very large number, with a computational complexity of $\Theta(N^m)$. For the systems that are being modeled and studied, explicitly solving this combinatorial optimization problem is infeasible. Thus, we need to devise a computationally efficient method for determining the $m$ best sensing regions that approximates the combinatorial optimization solution. 

\subsection{Online adaptation of models}\label{subsec:online-adpt}
In this work, the data can be assimilated at varying frequencies, allowing for single updates, with the acquisition of data at a single time, or batch updates, with the acquisition of data over some period of time. This flexibility reduces connectivity requirements. For example, robots can explore their assigned optimal sensing locations for some time without maintaining connectivity with the rest of the robots, to return after some period of time, share their acquired data, and update the model according to the online model adaptation algorithm.

The assimilation of newly acquired data into the models after a sensing period is a fundamental component of modeling a dynamic environment. In trying to model a dynamic process, new sensing data must be assimilated into the existing model to adequately capture the changing environment. Given the dimensionality of the data and the distributed nature of collecting sensing measurements, it is infeasible for all robots to keep the full time series data and global models on-board. The techniques for online assimilation of the data to the existing model may also depend on the duration of the process and computational limitations of the vehicles. The key challenge in this framework is determining which data and features are needed in updating and maintaining models of the environment.

Different assimilation strategies may be used for different types of dynamics and environments. Some environmental processes require long-term monitoring of extremely high-dimensional spaces. Alternatively, other types of environmental processes are quickly time varying processes with lower dimensionality. Thus, various methodologies with diverse computational and algorithmic constraints are required to efficiently adapt the models. This is addressed by selecting different model adaptation strategies for the type of monitoring. For example, the technique for extremely high-dimensional spaces has to address memory and computational efficiency. Alternatively, the technique for quickly time varying processes has to provide greater emphasis on newly acquired data. Furthermore, the adaptation techniques are constrained by the choice of model, as in the online data assimilation mechanisms must be consistent with the modeling and optimal sensing strategies devised.

\section{Proposed approach}
\label{sec:proposed-approach}
\begin{figure}
\includegraphics[width=0.45\textwidth]{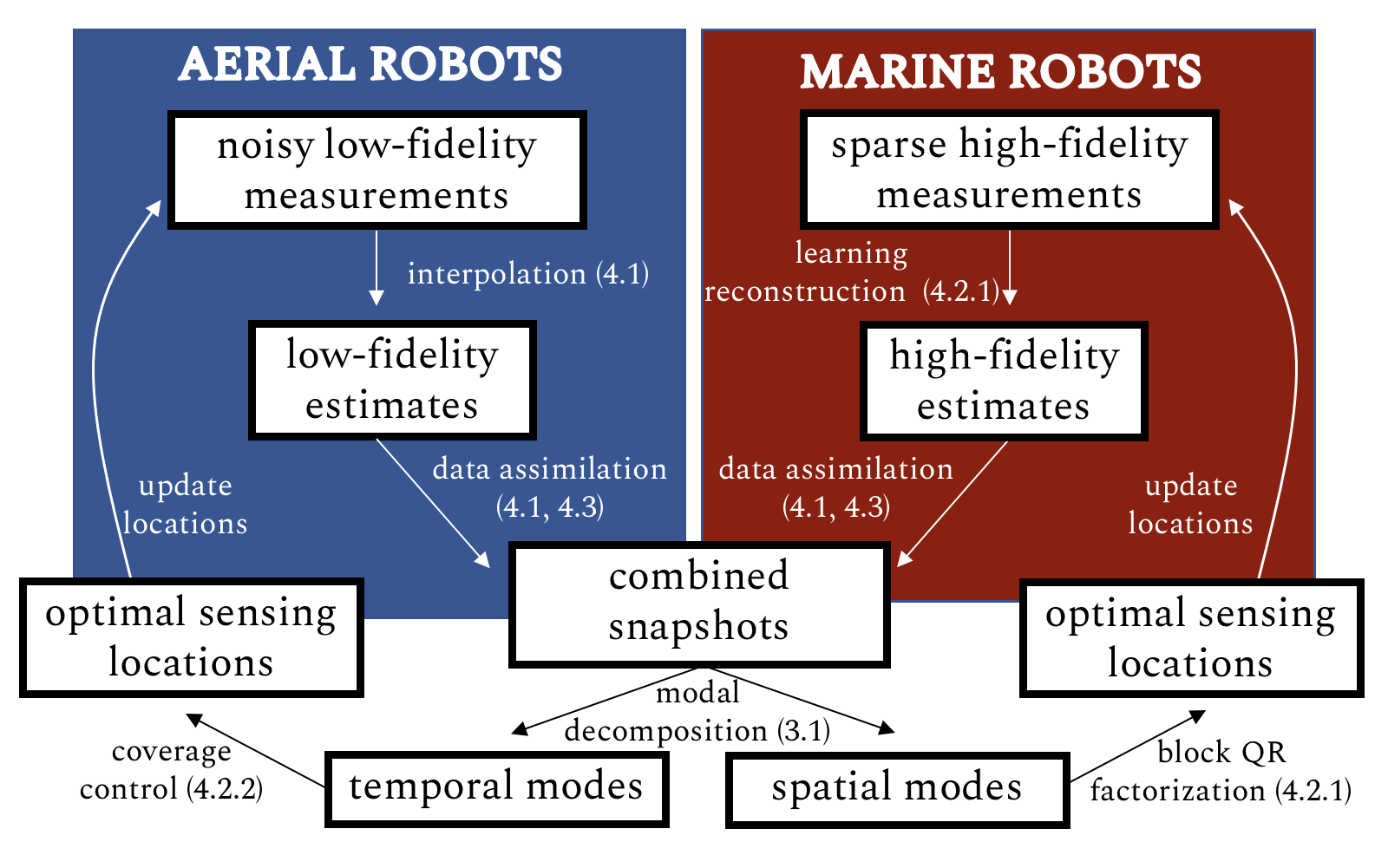}
\caption{Schematic of proposed framework with corresponding section numbers. Noisy low fidelity measurements from aerial vehicles are upsampled and sparse high-fidelity sensor measurements from marine vehicles are used to extrapolate the full state. Estimates are combined and used as input to the modal decomposition algorithm that provides decoupled temporal and spatial characteristics. Characteristics are used independently to determine the optimal sensing locations for data collection and eventual data assimilation}
\label{fig:framework-schematic} 
\end{figure}

The proposed framework, demonstrated in Fig. \ref{fig:framework-schematic}, solves the model estimation and uncertainty quantification through interpolation and extrapolation techniques that allow for the combination of various data sources and outputs temporal and spatial characteristics of the environment. The uncertainty-reducing task assignment is solved efficiently by leveraging these characteristics from the modal decomposition. Finally, the solution for online adaptation of the model relies on the properties of the modes itself. 

\subsection{Reconciling spatial and temporal scales for model construction}
\begin{figure*}
\includegraphics[width=0.99\textwidth]{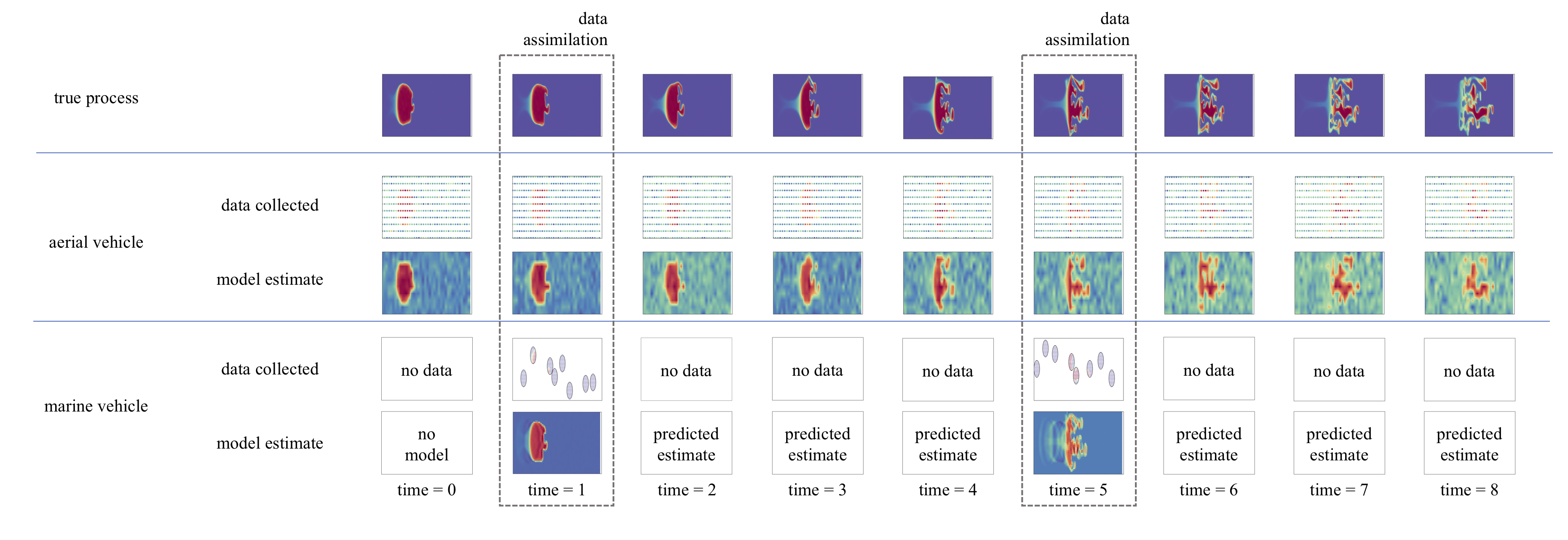}
\caption{Overview of data collected by heterogeneous multi-robot team used to produce low- and high-fidelity estimates of field. Low-fidelity measurements from the aerial vehicles are used to create low-fidelity estimates of the true process as shown at $\text{time} = 4$. High-fidelity sensor measurements, which are not always available due to sensing and computational limits, are collected at specific sensing locations by the marine vehicles, as shown by the circular regions at $\text{time} = 1$. A more detailed combined model is constructed using aerial vehicle model estimates and sparse sensor measurements from the marine vehicle. This model can be used for prediction when there is no marine vehicle data available. Predictions and new sensor measurements are assimilated into the model. The combined model is able to capture more interesting patterns through the strategic placement of the sensors, as seen at $\text{time} = 5$}
\label{fig:resolution}
\end{figure*}
In order to determine the informativeness of the points in the workspace, first we must define a method for unifying the measurements from disparate sources. This requires combining data into one cohesive model, while maintaining the independence of data sources. Though the data from different sources is attempting to model and predict the same phenomena, they come at various temporal and spatial resolutions. In order to resolve the various temporal rates and spatial resolutions, we assure that the data collected at faster rates but with lower spatial resolutions are interpolated to the same resolution as the data collected at higher spatial resolution. This higher spatial resolution data is being collected at a lower rate. Additionally, data collected at higher spatial resolution but with sparse measurement, as in the case where robots can densely sample but only at limited sensing location, is estimated at high spatial resolution across the entire discretization of the workspace. This is illustrated in Fig. \ref{fig:resolution}.

\begin{figure*}[ht!]
\centering
\subfloat[ ]
   {\includegraphics[width=1\linewidth]{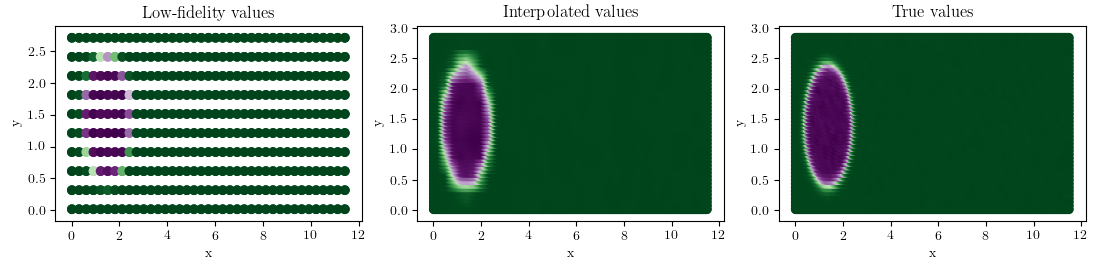}\label{interp-early}}

\subfloat[ ]
   {\includegraphics[width=1\linewidth]{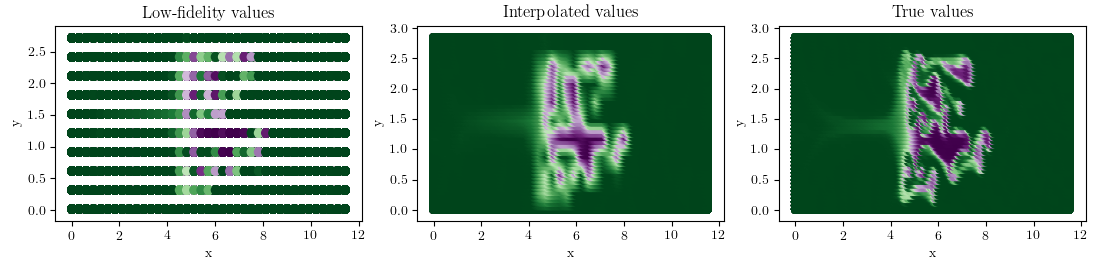}\label{interp-late}} 
   \caption[Sampling of low-fidelity samples to higher spatial resolution.]{Sampling of low-fidelity samples to higher spatial resolution. The original field has measurements on a $96 \times 384$ spatial grid. (a) At time = $1.004$ secs, low-fidelity measurements on a $10 \times 39$ grid are upsampled to same spatial scale as high-fidelity measurement on a $48 \times 192$ grid and closely agree with the true field. (b) At time = $97.22$ secs, upsampling of low-fidelity measurements to same spatial scale as high-fidelity measurement fail to capture relevant features of true field} \label{interp-change}
\end{figure*}

In the model-dependent optimal sensor allocation component, the data collected by robots is combined and then represented using DMD \citep{Schmid2010DynamicData}. For completeness, we review some of the key ideas behind DMD, first presented in Sect. \ref{subsec:background}, again here. DMD learns a low-dimensional model that contains a set of modes, where each mode is associated with a fixed oscillation frequency and decay or growth rate, from the collected data. Using DMD allows for the observation of oscillations in dynamic processes with both short and long time scales \citep{Tu2013OnApplications}. Additionally, the output model from DMD has physically interpretable meaning and can be used to determine optimal sensing locations in space \citep{Mezic2005, Manohar2018Data-DrivenPatterns, Manohar2019OptimizedDynamics}. However, given the inherent multiresolution nature of the data collected by the heterogeneous robots, the data must first be resolved into a cohesive model. There are various ways to do this, including sequential DMD, non-sequential DMD, \citep{Tu2013OnApplications}. In this paper, we use standard interpolation techniques to resolve the data to a uniformly sampled model.

In Sects. \ref{sec:simulation-analysis} and \ref{sec:experimental-validation}, the focus is on marine vehicles sampling with high spatial resolution at sparse locations, and aerial vehicles sampling at low spatial resolution but with a faster temporal rate. While the proposed approach holds true for this scenario, we argue that this framework is also applicable to heterogeneous teams in general. The approach provides a model and prediction scheme irrespective of the natures of the data sources provided.

We propose combining the measurements at varying temporal and spatial scales of the process into one compact model that provides the dominant spatial and temporal characteristics of the process. Initially, models of two types are constructed: high-fidelity models with sparse measurements from marine robots, as they collect precise measurements of the environment but only within sensing regions, and low-fidelity models with abundant measurements from aerial robots, as they survey large areas but cannot obtain precise samples. High-fidelity models provide information for areas of the dynamic process to allow for locally optimal estimates of the environment. Low-fidelity models characterize high-level descriptions of the dynamic process. While we use aerial and marine robots, more generally, multi-fidelity models encapsulate heterogeneity from various sources, such as from robots with different mobility or sensor types. Models from aerial robots are processed at a much higher frequency, due to their increased computation power, while models from marine robots are processed at lower frequency due to the fact that the data comes in much slower and must be processed to estimate the full field. The aerial vehicles maintain a coverage formation and upsample their low-fidelity measurements to the same spatial resolution as the high-fidelity measurement \citep{Cortes2004CoverageNetworks}.

Initially, for some time $T + 1$, the low-fidelity measurements from the aerial vehicles are collected. The aerial vehicles assume their positions using coverage control and assume a uniform density function across the space. The low-fidelity measurements are upsampled using bilinear interpolation, such that spatial resolution of the interpolated data is the same as the candidate high-fidelity data. In this scenario, standard interpolation techniques are easy to implement, computationally efficient, and perform just as well as other complicated learning techniques, such as neural networks. In our studies, we use standard interpolation as there was not a statistical difference in estimation when compared to other learning techniques and standard interpolation does not require prior training and historical knowledge. However, this method is general enough to use alternative estimation techniques, so long as the dimensions are compatible with the high-fidelity estimates. While there is obviously some relationship between the various fidelity sensor measurements, we do not take this into account in the upsampling as to account for the independence in sensor measurements from disparate sensing sources. With this technique, we capture the coarse spatial features of the environment but at a much higher frequency. Unsurprisingly, for simple spatial features, this approach is successful, as shown in Fig. \ref{interp-change} \subref{interp-early}. In these instances, the interpolated values closely match the true values of the field. However, for more complex features, the simple interpolation method is alone not sufficient Fig. \ref{interp-change} \subref{interp-late}. The interpolated do not match the true values of the field. 

Given a dataset that is now characterized by a fast temporal rate and high spatial resolution, we can use dimensionality reduction techniques to i) build a cohesive model of the system that can be used for estimation of missing sensor values, and ii) determining the relative importance of points in the workspace to best estimate the dynamic process. Specifically, given a combined data set $\mathbf{X}$, we can use the DMD analysis described in Sect. \ref{sec:problem-statement} to compute a reduced order model. As given by the reconstruction equation, Eq. \eqref{eq:dmd-reconstructed}, any state of the dynamical system at time $t$, $\mathbf{x}(t)$, can be approximated using the DMD modes and eigenvalues as $\hat{\mathbf{x}}(t) = \mathbf{\Phi}\mathbf{\Lambda}^{\top} \bm{\alpha}$. Note, that for the DMD modes, $\mathbf{\Phi} \in \mathbb{R}^{N \times r}$, where $N$ is the dimension of the discretization of the workspace. This means that the row $i$ of the matrices $\mathbf{\Phi}$ and $\mathbf{\Lambda}$ corresponds to a specific point $p_i$ from the discretization of the workspace. In Sect. \ref{sec:problem-statement}, we discussed how $\mathbf{\Phi}$ captures the key spatial modes and $\mathbf{\Lambda}$ captures the temporal dynamics of the spatial modes. We will use the spatial information in the approximation relationship, captured by the matrix $\mathbf{\Phi}$, to reconstruct the system using the high-fidelity sensor measurements from the marine robots. Similarly, this relationship will also be leveraged in determining the optimal sensing locations for the marine robots. This will be discussed in further detail in the following section. Alternatively, we will use the temporal information information, captured by the matrix $\mathbf{\Lambda}$ to construct a density function for a coverage formation for the aerial vehicles \citep{Cortes2004CoverageNetworks}. As the components of $\mathbf{\Lambda}$ represent either the rate of growth or decay and frequency of oscillation of the corresponding spatial modes, we use this to emphasize areas of either high growth or high decay in the density function in an attempt to capture some of the quickly time varying features that may be missed by the marine robots.

\subsection{Fast task allocation for optimal sensing locations}
Given the spatial and temporal characteristics of the environment, we can use these to inform us of the next sensing locations for the robots. For robots sampling at select sensing regions and receiving high-fidelity data, the optimization problem in Eq. \eqref{eq:opt} can be solved using the spatial characteristics (Sect. \ref{subsubsec:block-qr}). To determine optimal sensing locations for robots receiving low-fidelity measurements over larger regions at a faster rate we rely on standard coverage control techniques using the temporal characteristics (Sect. \ref{subsubsec:coverage}).

\subsubsection{Block QR pivoting using spatial characteristics} \label{subsubsec:block-qr}
To begin, $m$ robots collecting high-fidelity sensing measurements are initially deployed to random positions. Given that robots taking high spatial resolution measurements can only be located at $m$ points, the locations needed to be selected such that the informativeness of the sensing data collected is maximized. To reiterate, using the definition of sensing regions $S_i$ provided in Eq. \eqref{eq:sensing-region}, $L$ is a set containing the elements within the sets $S_1, S_2, \dots, S_m$ that are arbitrary sensing regions, such that for any two sets $S_i$ and $S_j$ with elements in $L$, $S_i \cap S_j = \varnothing$. Let $\mathbf{x}_{L}$ contain the states only at the locations specified by the set $L$, meaning $\mathbf{x}_{L}$ is just a vector of sensor measurements. Furthermore, let $\mathbf{C}_{L}$ be the observation matrix, such that $\forall p_i \in L$, $
{\mathbf{C}_L}_{ii} = 1$ and for all other entries, ${\mathbf{C}_L}_{jk} = 0$. 

For sensor measurements with additive Gaussian white noise, $\eta \sim \mathcal{N}(0, \sigma^2)$, the relationship between the full state, $\mathbf{x}$, and the available sensor measurements, $\mathbf{x}_L$ are defined as

\begin{align}
    \mathbf{x}_L &= \mathbf{C}_L\mathbf{x} + \eta \\
                 &= \mathbf{C}_L \mathbf{\Phi}\mathbf{\Lambda}^{\top} \bm{\alpha} + \eta.
\end{align}

When the full state is unknown, as in the case with collecting high-fidelity sensor measurements at sparse locations, the full state must be predicted from the observed state. In this case, the matrix $\mathbf{C}$ is fixed by our choice of $L$, the sensing locations determined optimal as in Eq. \eqref{eq:opt}. Then, the optimal least-squares estimate for the full state is determined by approximating the unknown quantity $\mathbf{a}(t) = \mathbf{\Lambda}^{\top} \bm{\alpha}$ in Eq. \eqref{eq:dmd-reconstructed}. The optimal least-squares then gives us $\mathbf{a}(t) \approx (\mathbf{C}_L \mathbf{\Phi})^{\dagger} \mathbf{x}_L = \mathbf{\hat{a}}(t)$. Substituting $\mathbf{\hat{a}}(t)$ back into the DMD estimation from Eq. \eqref{eq:dmd-reconstructed} gives the reconstructed full state $\mathbf{\hat{x}}$ from sensor measurements $\mathbf{x}_L$ as
\begin{equation} \label{eq:gappy-pod}
    \mathbf{\hat{x}} \approx \mathbf{\Phi} (\mathbf{C}_L \mathbf{\Phi})^{\dagger} \mathbf{x}_L.
\end{equation} 

This procedure is known as the gappy POD method \citep{Everson1995Karhunen--LoeveData} and has been well-studied in using sparse measurements to reconstruct a process of interest from basis functions \citep{Willcox2006UnsteadyDecomposition, Manohar2019OptimizedDynamics}. 

There is the central question of how to select the set $L$, and consequently the matrix $\mathbf{C}_L$, such that it fulfills the objective of minimizing the least-squares approximation error between an estimated $\mathbf{a}(t) - \mathbf{\hat{a}}(t)$.  One metric for estimating the least-squares approximation error is the error covariance, $\mathbf{K}$, \citep{Joshi2009SensorOptimization, Manohar2018Data-DrivenPatterns}, where 
\begin{equation}
    \mathbf{K} = Var(\mathbf{a} - \mathbf{\hat{a}}) = \sigma^2\left[\left(\mathbf{C}_L \mathbf{\Phi} \right)^\dagger \mathbf{C}_L \mathbf{\Phi} \right]^{-1}.
\end{equation}
The error covariance $\mathbf{K}$ describes the $\rho$-confidence ellipsoid, $\varepsilon_{\rho}$, that contains the least squares error $\mathbf{a} - \mathbf{\hat{a}}$ with probability $\rho$ \citep{Joshi2009SensorOptimization}. Intuitively, this means that the ellipsoid $\varepsilon_{\rho}$ contains all the vectors $\mathbf{\hat{a}}$ that could be $\mathbf{a}$ with confidence $\rho$. In this sense, we are seeking to minimize the volume of the ellipsoid, defined as vol($\varepsilon_\rho$) = $\delta_{\rho, r}$det$\mathbf{K}^{1/2}$.

When picking the $p_i \in L$ optimal point locations from which sensor measurements can be obtained, with $\mathbf{C}_L$ determined by $L$, this optimization problem can be formulated as 
\begin{align} \label{eq:l-star-opt}
    L^{\star} = \argmax_{L} \text{det}\left[\left(\mathbf{C}_L \mathbf{\Phi} \right)^\dagger \left(\mathbf{C}_L \mathbf{\Phi}\right) \right].
\end{align}
We first note that the minimization of the determinant of $\mathbf{K}$, which minimizes the volume of the error ellipsoid, is equivalent maximization of the determinant to the inverse $\left[\left(\mathbf{C}_L \mathbf{\Phi} \right)^\dagger \left(\mathbf{C}_L \mathbf{\Phi}\right) \right]$. The optimal solution to this requires a combinatorial search of $\Theta(N^m)$ over the space of high-dimension, $N$, and $m$ candidate sensing locations. However, the solution to Eq. (\ref{eq:l-star-opt}) can be found via an extremely efficient greedy optimization method using a modified the pivoted QR factorization of matrices \citep{Manohar2018Data-DrivenPatterns, Manohar2019OptimizedDynamics}. The pivoted QR factorization is used to maximize the volume of successive submatrices, meaning the absolute value of the determinant is also being maximized.

We first describe the traditional QR factorization method used for optimal design, and then discuss our novel modifications that allow the method to be generalized to our purposes. The traditional QR factorization differs from our method in two regards: i) it does not explicitly offer a weighting scheme to evaluate the quality of individual measurements, and ii) it does not take into account sensing radius. In general, optimal sensor placement methods focus on sensing at a fixed number of  disjoint points. Alternatively, in this approach we have robots with a finite sensing radius over a set of points, as in Eq. \eqref{eq:sensing-region}. Thus, the optimization problem is jointly finding the best set of points over all available robots to collect sensor measurements, as in Eq. \eqref{eq:opt}. In the modifications presented, we address both of these issues in a computationally efficient way.

First, we present the traditional QR algorithm for optimal design, as shown in Alg. \ref{alg:pivot-qr}. Then, we will present our pivoted QR factorization algorithm with modifications made for selecting $m$ sensor locations based on sensing radius, $k$, as shown in Alg. \ref{alg:pivot-qr-sensing}. We will show that these modifications admit a greedy solution that maximizes the absolute value of the determinant, in Eq. \eqref{eq:l-star-opt}, while taking into account the sensing constraints of the robots.

\begin{figure*}
\centering
\subfloat[ ]{
  \includegraphics[width=42mm]{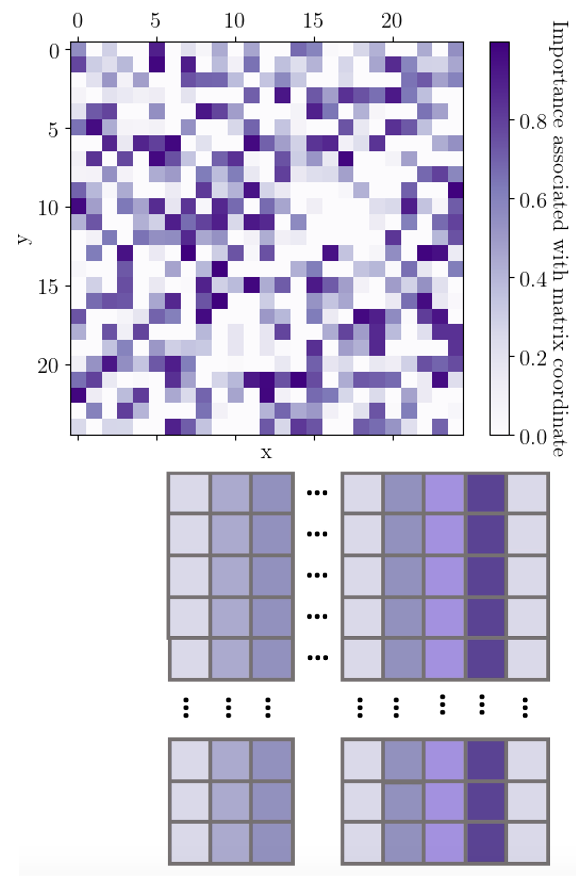}
}
\subfloat[ ]{
  \includegraphics[width=42mm]{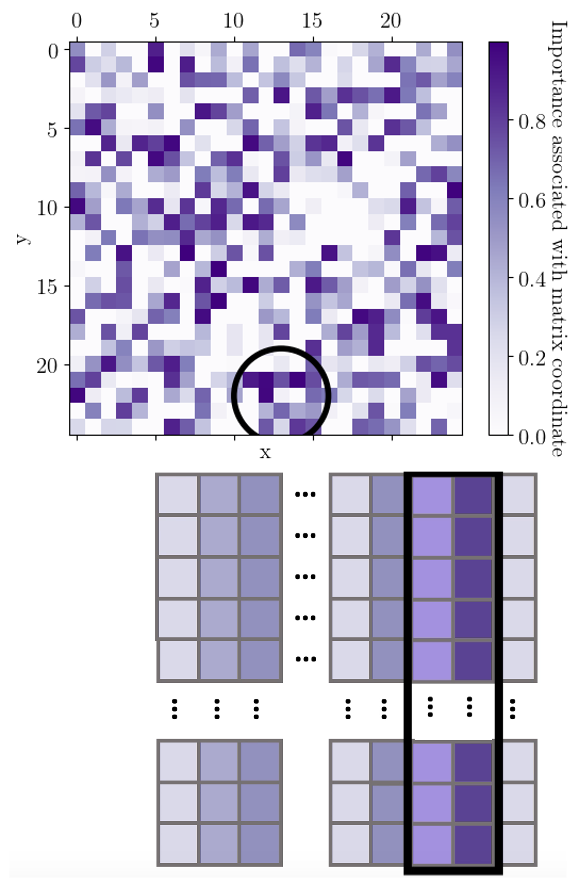}
}
\subfloat[ ]{
  \includegraphics[width=42mm]{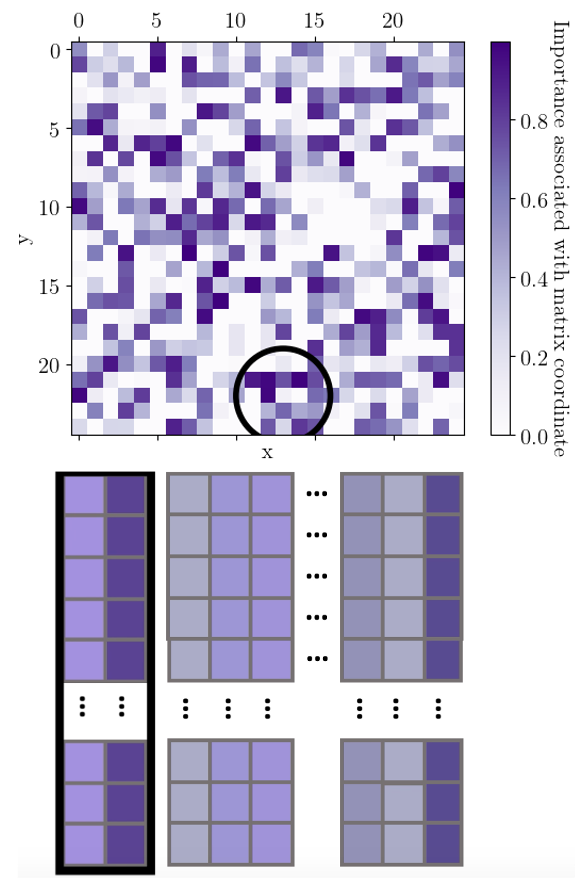}
}
\subfloat[ ]{
  \includegraphics[width=42mm]{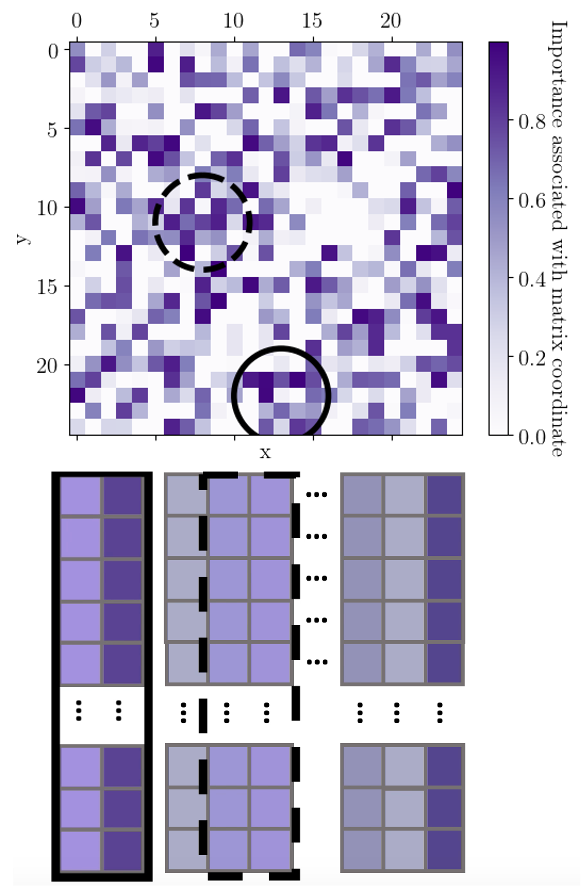}
}
\caption{Block pivoted QR factorization for optimal sensor location. (a) The locations in the environment have a informativeness associated with collected sensing at that point, where darker colors correspond to greater informativeness. This information is capture in a projection matrix, with columns of the corresponding to individual sensing locations. (b) The maximally informative sensing location is selected as the the solid line circle according to the maximization weighting scheme that selects the associated optimal columns. (c) The matrix columns are pivoted and weights of the remaining locations are adjusted by removing its orthogonal projection. (d) A new maximally informative sensing location is selected as the dash line circle and the process is repeated}\label{fig:qr-pivot-block}
\end{figure*}

In the traditional QR algorithm, QR factorization with column pivoting gives us the decomposition $\mathbf{\Phi} \mathbf{\Phi}^{\top} \mathbf{C}^{\top} = \mathbf{QR}$. The pivoting provides an approximate solution for the optimization problem in Eq. \eqref{eq:l-star-opt}. This procedure is known as submatrix volume maximization since the matrix volume is the absolute value of the determinant and at each iteration of the algorithm, the absolute determinant is greedily maximized. In the QR column pivoting procedure, a pivot column with the maximal 2-norm is selected that corresponds to the optimal sensing point with respect to minimizing the error ellipsoid. The orthogonal projection of all of the remaining columns onto this column is subtracted from the remaining columns.

The method of using blocks within the pivoted QR algorithm admits a valid factorization of the desired $\mathbf{\Phi}\mathbf{\Phi}^\top$ that allows us to greedily optimize the absolute value of the desired matrix for a specified sensing radius $k$. This blocked pivoted QR is demonstrated in Fig. \ref{fig:qr-pivot-block}. In this scenario, we can perform our block version of the pivoted QR factorization of $\mathbf{\Phi}\mathbf{\Phi}^{\top}$ since these values will coincide with the groups of singular values of the $\left[\left(\mathbf{C} \mathbf{\Phi} \right)^{\top} \mathbf{C} \mathbf{\Phi} \right]^{-1}$ \citep{Manohar2019OptimizedDynamics}. Pivoted QR factorization of matrices decomposes a matrix $\mathbf{M}$ into a unitary matrix $\mathbf{Q}$ and an upper-triangular matrix $\mathbf{R}$ such that $\mathbf{\Phi} \mathbf{\Phi}^{\top} \mathbf{C}^{\top} = \mathbf{QR}$. Using the properties of determinants, we have that $|\text{det} \mathbf{\Phi} \mathbf{\Phi}^{\top} \mathbf{C}^{\top}| = |\text{det}\mathbf{Q}||\text{det}\mathbf{R}| = \prod_i |\mathbf{R}_{ii}|$. The QR pivoting enforces the diagonal dominant structure on the diagonal entries of the matrix $\mathbf{R}$ such that $\sigma_i^2 = |\mathbf{R}_{ii}|^2 \geq \sum_{j=i}^c|\mathbf{R}_{jc}|, \quad 1 \leq i \leq c \leq T + 1$. This then maximizes the absolute value of the determinant. 

The block pivoted QR factorization evaluates all of the sets of columns that correspond to candidate, potentially non-adjacent sensing locations from the matrix $\mathbf{\Phi}\mathbf{\Phi}^\top$. The candidate sensing locations, $\mathcal{C}$, are determined by first enumerating the set of all points in each possible circle of radius $k$, corresponding to the sensing radius, within the workspace $\mathcal{W}$. These candidate circles within $\mathcal{W}$ are mapped to the columns corresponding to spatial locations in the matrix. The optimal sensing location is the set of columns within the matrix $\mathbf{\Phi}\mathbf{\Phi}^\top$ that correspond to the spatial regions $S^*$ where the sum of the column 2-norms $\sum_{p \in S_i}{||(\mathbf{\Phi}\mathbf{\Phi}^\top)_{p}||_2}$ is maximized. The algorithm then selects this optimal sensing location from the matrix and pivots it to the beginning of the matrix, removing its orthogonal projection from the remaining columns of matrix $\mathbf{\Phi}\mathbf{\Phi}^\top$. The candidate sensing locations are pruned such that any sensing locations with points in the selected optimal sensing location are discarded. This process is repeated the until all optimal sensing locations have been found. 

\begin{algorithm}
    \caption{Pivoted QR factorization algorithm for determining optimal point measurements}
    \label{alg:pivot-qr}
    \begin{algorithmic}[1]
    \Procedure{Pivoted QR Factorization}{$\mathbf{\Phi}\mathbf{\Phi}^\top$, $m$}
    \State $P \gets []$ \Comment{Set of selected point measurements}
    \State $W \gets []$ \Comment{Weights corresponding to points}
    \For{$i = 1, \dots, k$}
        \State $p = argmax_{j \not\in P}||(\mathbf{\Phi}\mathbf{\Phi}^\top)_j||_2$ \Comment{$j$-th column}
        \State $P \gets [P \quad p]$
        \State Determine Householder matrix $H$
        \State $\mathbf{\Phi}\mathbf{\Phi}^\top \leftarrow diag(I, H) \mathbf{\Phi}\mathbf{\Phi}^\top$ \Comment{\parbox[t]{.37\linewidth}{Remove orthogonal projection of $(\mathbf{\Phi}\mathbf{\Phi}^\top)_s$ from all columns}}
    \EndFor
    \State \textbf{return} $P, W$
    \EndProcedure
    \end{algorithmic}
\end{algorithm}

\begin{algorithm}
    \caption{Pivoted QR factorization algorithm for determining optimal sensor locations and their corresponding weights}
    \label{alg:pivot-qr-sensing}
    \begin{algorithmic}[1]
    \Procedure{Pivoted QR Factorization}{$\mathbf{\Phi}\mathbf{\Phi}^\top$, $m$, $\mathcal{W}$, $k$}
    \State $\mathcal{C} \gets$ \Call{create candidate sensing location}{$\mathcal{W}$, $r$}
    \State $L^{*} \gets []$ \Comment{Set of optimal sensing locations}
    \State $W \gets []$ \Comment{Weights corresponding to points}
    \For{$i = 1, \dots, m$}
        \State $S^{*} = argmax_{S_i \in S, S_i \not\in L^{*}}\sum_{p \in S_i}{||(\mathbf{\Phi}\mathbf{\Phi}^\top)_{p}||_2}$ 
        \State $L^{*} \gets [L^{*} \quad S^{*}]$
        \State $\mathcal{C} \gets \Call{remove overlapping regions}{\mathcal{W}, r, S^{*}}$
        \State $w = \frac{\norm{(\mathbf{\Phi}\mathbf{\Phi}^\top)_{S^{*}}}_2}{\sum_{i=0}^{n}\norm{\mathbf{\Phi}\mathbf{\Phi}^\top}_2}$ 
        \State $W \gets [W \quad w]$
        \State Determine Householder matrix $H$
        \State $\mathbf{\Phi}\mathbf{\Phi}^\top \leftarrow diag(I, H) \mathbf{\Phi}\mathbf{\Phi}^\top$
        \Comment{\parbox[t]{.37\linewidth}{Remove orthogonal projection of $(\mathbf{\Phi}\mathbf{\Phi}^\top)_{S^{*}}$ from all columns}}
    \EndFor
    \State \textbf{return} $S, W$
    \EndProcedure
    \end{algorithmic}
\end{algorithm}

\subsubsection{Coverage control using temporal characteristics} \label{subsubsec:coverage}
For robots with low spatial resolution measurements collected over time, a coverage control technique can be used to determine the placement of the robots.

In coverage control, the workspace $\mathcal{W}$ is partitioned into Voronoi regions, such that each robot $j$ located at point $p_j$ is sensing the points in the Voronoi region $V_j$ closest to its position, as in 
\begin{equation}\label{eq:voronoi}
    V_j = \{q \in \mathcal{W} \, | \,  \norm{q - p_j} \leq \norm{q - p_i}, i \neq j\},
\end{equation}
where $\mathcal{I}$ is the index set of all the points in the workspace. Let $R$ be the set of locations of all $m'$ robots for this sensing modality.

The coverage control technique considers minimizing the function
\begin{equation} \label{eq:coverage}
    \mathcal{H}(R) = \sum\limits_{j=1}^{m'} \int_{V_j} f\left(\norm{q- p_j}\right)\phi(q)dq.
\end{equation}

In coverage control, the density function $\phi$ is a measure of information or probability that some event takes place. In this scenario, we can use the temporal characteristics $\mathbf{\Lambda}$ as the density function $\phi$ in Eq. \eqref{eq:coverage} and use standard implements for coverage control techniques to determine the optimal locations for these robots. The minimization of the function Eq. \eqref{eq:coverage} can be solved using centroidal Voronoi partitions. These partitions correspond to the optimal partition of the space according to the density function such that the best coverage is achieved. We refer the interested reader to the work by \cite{Cortes2004CoverageNetworks} for full details on coverage control methods and implementation. The optimal sensing locations of the vehicles collecting low-fidelity sensor measurements are determined by partitioning the region according to the centers of the centroidal Voronoi partitions.

\subsection{Online adaptation of spatial and temporal features} \label{subsec:online-adaptation}
We proposed two methods to allow for the online adaptation of the DMD modes and eigenvalues. First, we propose a general method that works on large dimensional systems and provides a robust approximation of the DMD operator. However, this framework requires the retention of all of the snapshots collected. The second method presented allows for fast computations and does not require the retention of all of the snapshots collected, but imposes restrictions on the workspace. The latter framework is amenable to long-term monitoring of environmental phenomena.

In both procedures, we want to incorporate updates to the eigendecomposition of the operator $\mathbf{A}$ after the acquisition of new data $\mathbf{x_{T+1}}$, $\mathbf{x_{T+2}}, \dots \mathbf{x_{T+1+\tau}}$. Define $\mathbf{X_{new}}$ and $\mathbf{Y_{new}}$ as
\begin{gather}
    \mathbf{X_{new}} = [\mathbf{x_{T+1}} \, \mathbf{x_{T+2}} \, \dots \, \mathbf{x_{T+\tau}}] \\
    \mathbf{Y_{new}} = [\mathbf{x_{T+2}} \, \mathbf{x_{T+3}} \, \dots \, \mathbf{x_{T+1+\tau}}]
\end{gather}

with the combined data in matrices 
\begin{gather}
    \mathbf{X^{\prime}} = [\mathbf{X} \quad \mathbf{X_{new}}] \\
    \mathbf{Y^{\prime}} = [\mathbf{Y} \quad \mathbf{Y_{new}}].
\end{gather}

Then, for the online adaptation, the desired decomposition is for the operator $\mathbf{A}'$ for the equation $\mathbf{Y}' = \mathbf{A}'\mathbf{X}'$.

\subsubsection{\textbf{Generalized} adaptation for varying spatial and temporal scale modeling}
This method allows for adapting the eigenmodes and eigenvectors of $\mathbf{A}$ to $\mathbf{A}'$ in the presence of new data collected from a variety of systems. This method is referred to as the ``general'' method. While this method works with different spatial and temporal scales, it requires storing the previous SVD of the data matrix $\mathbf{X}$ and all of the snapshots of the data. However, as in the case of the traditional DMD algorithm, this online adaptation scheme alleviates the need for directly computing $\mathbf{A}'$ and instead computes the eigendecomposition of an approximation of $\mathbf{A}'$. We generalize the techniques in \citep{Matsumoto2017On-the-flySquares}, which allow for single updates at a time, using the results from \citep{Brand2002IncrementalData} that allow for updating SVD computations with missing or incomplete data. We can formulate a new matrix $\mathbf{X^{\prime}}$ in terms of the SVD of the previous data matrix $\mathbf{X}$ from Eq. \eqref{eq:x-decomp} and the new data $\mathbf{X_{new}}$.

\begin{equation}
\begin{split}
\mathbf{X^{\prime}} & = [\mathbf{X} \quad \mathbf{X_{new}}] = [\mathbf{U} \bm{\Sigma} \mathbf{W}^{\top} \quad \mathbf{X_{new}}] \\
& = [\mathbf{U} \quad \mathbf{J}] \begin{pmatrix} \bm{\Sigma} & \mathbf{L} \\ 
\mathbf{0} & \mathbf{P} \end{pmatrix} \begin{pmatrix} \mathbf{W} & \mathbf{0} \\ 
\mathbf{0} & \mathbf{I} \end{pmatrix} ^ T \\
& = [\mathbf{U}(\mathbf{I} - \mathbf{UU}^{\top})/\mathbf{P}] \begin{pmatrix} \bm{\Sigma} & \mathbf{U}^{\top} \mathbf{X_{new}} \\ 
\mathbf{0} & \mathbf{P} \end{pmatrix} \begin{pmatrix} \mathbf{W} & \mathbf{0} \\ 
\mathbf{0} & \mathbf{I} \end{pmatrix} ^ T
\end{split}
 \end{equation}
where $\mathbf{L} = \mathbf{U}^{\top} \mathbf{X_{new}}$ is the projection of the new data onto the orthogonal basis $\mathbf{U}$, $\mathbf{P} = \mathbf{J}^{\top} \mathbf{H}$, $\mathbf{H} = \mathbf{U}(\mathbf{I} - \mathbf{UU}^{\top})\mathbf{X_{new}} = \mathbf{X_{new}} - \mathbf{UL}$ is the component of the new data orthogonal to the subspace spanned by $\mathbf{U}$, $\mathbf{J}$ is an orthogonal basis of $\mathbf{H}$ from the QR decomposition $\mathbf{JR} = \mathbf{H}$, and $\mathbf{K} = \mathbf{J}^{\top} \mathbf{H}$ be the projection of the new data onto the subspace orthogonal to $\mathbf{H}$.

Denote an intermediate matrix $\mathbf{Z}$ as

\begin{equation}
    \mathbf{Z} = \begin{pmatrix} \bm{\Sigma} & \mathbf{U}^{\top} \mathbf{X_{new}} \\ 
\mathbf{0} & \mathbf{P} \end{pmatrix}.
\end{equation}

We can then take the SVD of $\mathbf{Z} = \mathbf{U^{\prime}} \bm{\Sigma^{\prime}} \mathbf{W^{\prime T}}$

Finally, we can update $\mathbf{U}$, $\bm{\Sigma}$, $\mathbf{W}$.
\begin{gather}
    \mathbf{U} \leftarrow [\mathbf{U} \quad \mathbf{J}] \mathbf{U^{\prime}} \nonumber \\
    \bm{\Sigma} \leftarrow \bm{\Sigma^{\prime}} \nonumber \\
    \mathbf{W} \leftarrow \begin{pmatrix} \mathbf{W} & \mathbf{0} \nonumber \\ 
\mathbf{0} & \mathbf{I} \end{pmatrix} \mathbf{W^{\prime}} \nonumber
\end{gather}

With the updated $\mathbf{U}$, $\bm{\Sigma}$, $\mathbf{W}$, we can proceed with Eq. \eqref{eq:projected-A} using the updated SVD matrices and the updated $\mathbf{Y^{\prime}}$. This allows us to compute a low-dimensional approximation of the $\mathbf{A}'$ operator.

\subsubsection{Fast adaptation for \textbf{long-term} modeling of environmental phenomena with coarse spatial scale} Alternatively, this method allows for direct computation of the operator $\mathbf{A}'$ on data collected over an extended period of time. However, the spatial dimension of the environment being sampled should be small (less than 200). While the previous $\mathbf{A}$ matrix needs to be retained for updating, the snapshots do not. Instead, a compact, lower dimensional representation of the previous data is stored and updated. This method is referred to as the ``long-term'' method. We extend and generalize the results of \citep{Zhang2019OnlineSystems}, where the authors allow for estimating updates to the new $\mathbf{A}'$ matrix after the acquisition of a single new training datum. In this work, we allow for simultaneous batch updates to compute the new $\mathbf{A}'$ matrix.

Assuming that $\mathbf{X}$ is a matrix with full row rank, we can write the pseudoinverse of the matrix as 
\begin{equation}
    (\mathbf{X})^{\dagger} = (\mathbf{X})^{\top}(\mathbf{X} \mathbf{X}^{\top})^{-1}
\end{equation}

Thus, $\mathbf{A}$ can be written in terms of 
\begin{gather}
    \mathbf{A} = \mathbf{Y}(\mathbf{X})^{\top}(\mathbf{X} \mathbf{X}^{\top})^{-1} = \mathbf{Q}\mathbf{S} \\
    \mathbf{Q} = \mathbf{Y}(\mathbf{X})^{\top} \\
    \mathbf{S} = (\mathbf{X} \mathbf{X}^{\top})^{-1}.
\end{gather}

In order to calculate the updated $\mathbf{A}$, we need $\mathbf{A^{\prime}}$ = $\mathbf{Q^{\prime}} \mathbf{S^{\prime}}$, where we can write $\mathbf{Q^{\prime}}$ and $\mathbf{S^{\prime}}$ in terms of $\mathbf{Q}$, $\mathbf{S}$, and the new training data in $\mathbf{X_{new}}$ and $\mathbf{Y_{new}}$.

\begin{align}
    \mathbf{Q^{\prime}} & = [\mathbf{Y}  \quad \mathbf{Y_{new}}] [\mathbf{X} \quad \mathbf{X_{new}}]^{\top} \nonumber \\
    & = \mathbf{Y}\mathbf{X} +  \mathbf{Y_{new}} \mathbf{X_{new}^{\top}} \nonumber \\
    & = \mathbf{Q} \mathbf{S} + \mathbf{Y_{new}} \mathbf{X_{new}^{\top}} \label{q-n-k}\\
    (\mathbf{S^{\prime}})^{-1} &= [\mathbf{X} \nonumber \quad \mathbf{X_{new}}] [\mathbf{X} \quad \mathbf{X_{new}}]^{\top} \\
    & = \mathbf{X}\mathbf{X}^{\top} + +\mathbf{X_{new}}\mathbf{X_{new}}^{\top} \nonumber \\
    & = \mathbf{S^{-1}} +\mathbf{X_{new}}\mathbf{X_{new}}^{\top} \label{s-n-k}
\end{align}

The new $\mathbf{A^{\prime}}$ matrix is then 
\begin{align}
    \mathbf{A^{\prime}} &= \mathbf{Q^{\prime}} \mathbf{S^{\prime}} \nonumber \\ 
    &= (\mathbf{Q} \mathbf{S} + \mathbf{Y_{new}} \mathbf{X_{new}^{\top}}) (\mathbf{S}^{-1} + \mathbf{X_{new}}\mathbf{X_{new}^{\top}})^{-1}. \label{eq:expanded-a-prime}
\end{align}

From Eq. \eqref{eq:expanded-a-prime}, we see that $\mathbf{Q^{\prime}}$ is easily computed using the previously computed matrices $\mathbf{Q}$ and $\mathbf{S}$ and the newly acquired data. However, we still need to compute $\mathbf{S^{\prime}} = (\mathbf{S^{-1}} + \mathbf{X_{new}}\mathbf{X_{new}^{\top}})^{-1}$. To do so, we use the Woodbury matrix identity given by

\begin{equation}\label{eq:woodbury}
    (\mathbf{C} + \mathbf{DEF})^{-1} = \mathbf{C}^{-1} - \mathbf{C}^{-1}\mathbf{D}(\mathbf{E}^{-1} + \mathbf{FC}^{-1}\mathbf{D})^{-1}\mathbf{F}\mathbf{C}^{-1}.
\end{equation}

Let $\mathbf{C} = \mathbf{S}^{-1}, \mathbf{D} = \mathbf{X_{new}}, \mathbf{E} = \mathbf{I}, \mathbf{F} = \mathbf{X_{new}^{\top}}$ in Eq. \eqref{eq:woodbury},  we can rewrite $\mathbf{S^{\prime}}$ as
\begin{align} \label{eq:applied-woodbury}
    \mathbf{S^{\prime}} & = (\mathbf{S}^{-1} + \mathbf{X_{new}}\mathbf{X_{new}^{\top}})^{-1} \nonumber \\
    & = \mathbf{S} - \mathbf{S} \mathbf{X_{new}} (\mathbf{I} + \mathbf{X_{new}^{\top}} \mathbf{S} \mathbf{X_{new}})^{-1} \mathbf{X_{new}^{\top}} \mathbf{S}. 
\end{align}
Let $\mathbf{\Gamma} = (\mathbf{I} + \mathbf{X_{new}^{\top}} \mathbf{S} \mathbf{X_{new}})^{-1}$, combining Eqs. \eqref{eq:expanded-a-prime} and \eqref{eq:applied-woodbury}, we can arrive at a simplified expression for $\mathbf{A^{\prime}}$:
\begin{equation}
    \mathbf{A^{\prime}} = \mathbf{A} + (\mathbf{Y_{new}} - \mathbf{A} \mathbf{X_{new}})\mathbf{\Gamma} \mathbf{X_{new}^{\top}} \mathbf{S}.
\end{equation}

In this scenario, we also impose a weighting scheme to place greater importance on training data that has been collected more recently, which is helpful in the context of long-term monitoring where early training data may not be as useful. To weight samples with varying importance, we apply a forgetting factor $\gamma$ to past training data, receiving new expressions for $\mathbf{Q^{\prime}}$ and $\mathbf{S^{\prime}}$ given by
\begin{gather}
    \mathbf{Q^{\prime}} = \gamma \mathbf{Q} + \mathbf{Y_{new}} \mathbf{X_{new}^{\top}} \\
    (\mathbf{S^{\prime}})^{-1} = \gamma \mathbf{S^{-1}} +  \mathbf{X_{new}^{\top}} \mathbf{X_{new}}.
\end{gather}
Accounting for the weighting scheme, we modify the computation of $\mathbf{S^{\prime}}$ as 
\begin{equation}
    \mathbf{S^{\prime}} = \frac{1}{\gamma} * (\mathbf{S} - \mathbf{S} \mathbf{X_{new}} \mathbf{\Gamma} \mathbf{X_{new}^{\top}} \mathbf{S}).
\end{equation}

Thus, we can updated the operator $\mathbf{A}$ explicitly and use an eigenvalue decomposition to derive the DMD modes.

\section{Analysis of online algorithms}
\label{sec:simulation-analysis}
\begin{figure*}[ht!]
    \begin{minipage}{.33\textwidth}
    \centering
    \subfloat[]{\label{main:a}\includegraphics[width=57mm]{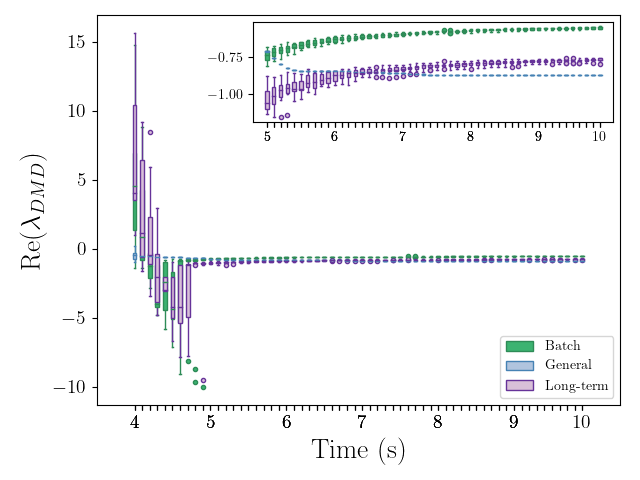}}
    \end{minipage}%
    \begin{minipage}{.33\textwidth}
    \centering
    \subfloat[]{\label{main:a}\includegraphics[width=57mm]{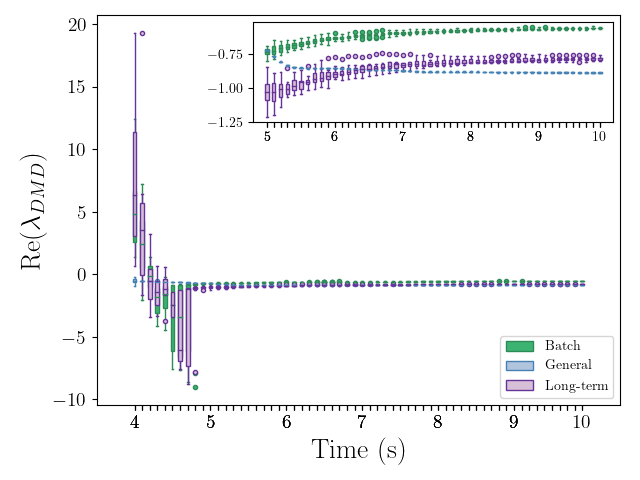}}
    \end{minipage}%
    \begin{minipage}{.33\textwidth}
    \centering
    \subfloat[]{\label{main:b}\includegraphics[width=57mm]{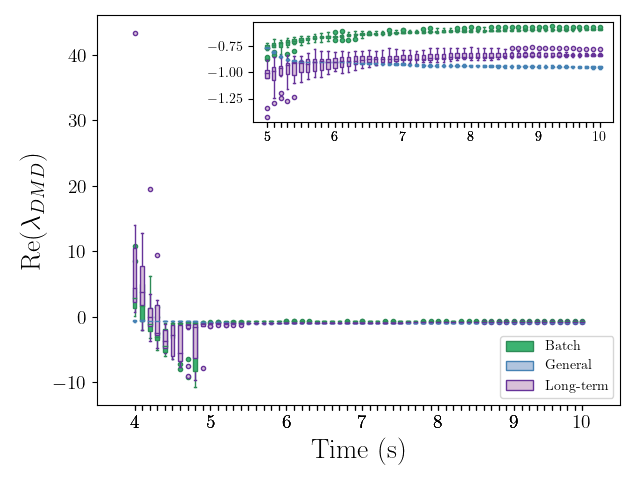}}
    \end{minipage}\par\medskip
    \begin{minipage}{.33\textwidth}
    \centering
    \subfloat[]{\label{main:a}\includegraphics[width=57mm]{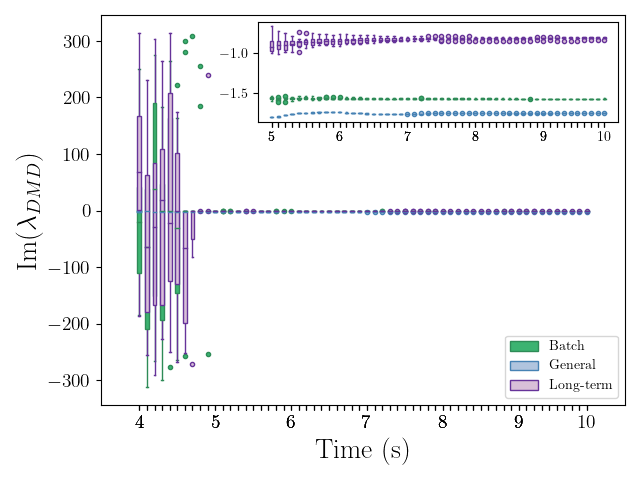}}
    \end{minipage}%
    \begin{minipage}{.33\textwidth}
    \centering
    \subfloat[]{\label{main:a}\includegraphics[width=57mm]{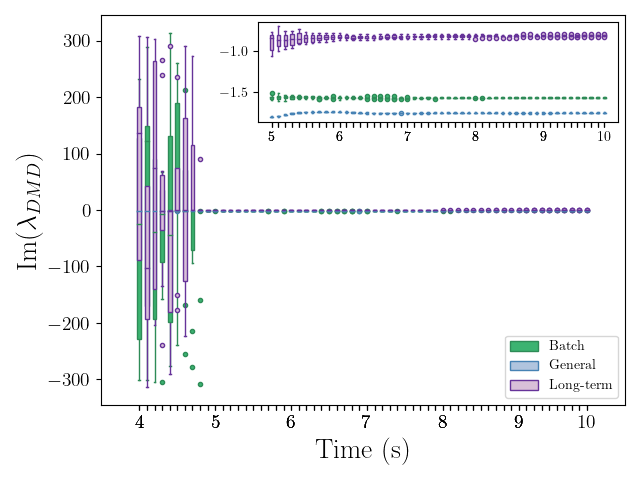}}
    \end{minipage}%
    \begin{minipage}{.33\textwidth}
    \centering
    \subfloat[]{\label{main:b}\includegraphics[width=57mm]{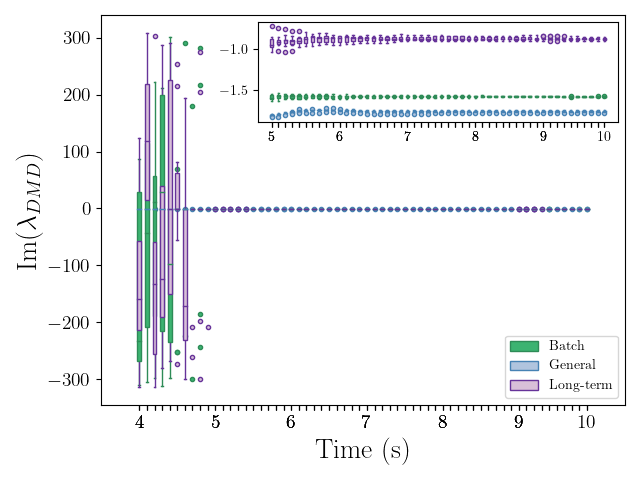}}
    \end{minipage}
    
    \caption{Comparison of various online DMD methods for computing eigenvalues. The data is collected in a $20 \times 20$ environment over $1000$ time steps from $0$ to $10$ seconds, where the online algorithms are initialized with data from $400$ time steps at $4$ seconds and updated every $10$ time steps. Various noise regimes are studied for additive white Gaussian noise with low, medium, and high variance. (a)-(c) Estimation of real part of eigenvalue of dynamical system, where true value is $-1$ where variance is $0.01$ in (a), $0.04$ in (b), and $0.1$ in (c). (d) - (f) Estimation of imaginary part of eigenvalue of dynamical system, where true value is $0$ where variance is $0.01$ in (d), $0.04$ in (e), and $0.1$ in (f)} \label{fig:eig-val-comparison}
\end{figure*}

\begin{figure}[ht!]
  \includegraphics[width=0.48\textwidth]{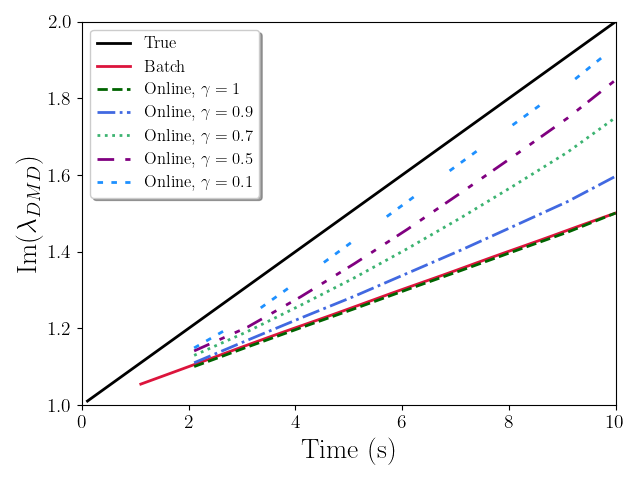}
\caption{Comparison of various weights using online data assimilation with DMD methods with batch DMD algorithm} \label{fig:online-weighting-comparison}
\end{figure}

\begin{table*}
\centering
\caption{Comparison of Execution Times for Online DMD Algorithms on Various Environments}
\begin{tabular}{|c|c||c|c|c|c|c|} \hline
\cline{1-6}
\multirow{4}{*}{\begin{tabular}{c}Environmental\\Parameters \end{tabular}} & \multirow{4}{*}{\begin{tabular}{c} Time \\ Dimension \\ $T$ \\ $\tau$ \end{tabular}} & $500$ & $500$ & $1000$ & $2000$ \\ \cline{2-6}
& & $10 \times 10$ & $10 \times 10$ & $20 \times 20$ & $20 \times 10$ \\ \cline{2-6}
& & 100 & 100& 400 & 200\\ \cline{2-6}
& & 10 & 100& 100 & 100 \\ \hhline{|======|} 
\multirow{3}{*}{\begin{tabular}{c}Online DMD  \\  Method (s) \end{tabular}} & \multirow{3}{*}{\begin{tabular}{c}\textbf{Batch} \\ \textbf{General} \\ \textbf{Long-term} \end{tabular}}  & 5.038 & 1.376 & 3.715 &  14.77 \\ \cline{2-6}
& & 0.557 &  0.975 & 1.710& 3.343\\ \cline{2-6}
& & 0.304 & 0.646 & 2.969 & 1.387 \\ \hhline{|======|}
\multirow{3}{*}{\begin{tabular}{c}With Eigen. \\ Computation (s) \end{tabular}} & \multirow{3}{*}{\begin{tabular}{c}\textbf{Batch} \\ \textbf{General} \\ \textbf{Long-term} \end{tabular}}  & 13.59 & 2.756 & 37.52 & 28.82 \\ \cline{2-6}
& & 0.537 & 1.950 & 3.411 & 4.266 \\ \cline{2-6}
& & 0.413 & 1.319 & 29.61 & 17.03 \\ \cline{1-6}
\end{tabular} \label{tab:exec-time}
\end{table*} 

In this section, we evaluate the proposed online algorithms, as described in Sect. \ref{subsec:online-adaptation}, in various environments with different time scales and resolutions. The two proposed approaches for online adaptation of the DMD modes and eigenvalues, general adaptation for varying spatial and temporal scale modeling and fast adaptation for long-term modeling of environmental phenomena with coarse spatial scale, have their respective advantages and drawbacks. As a basis for comparison, we will use the naive batch algorithm. At each update period $\tau$, the batch algorithm computes an estimate $\mathbf{A}_{batch}$ with $\mathbf{Y}' = \mathbf{A}_{batch} \mathbf{X}'$ using the pseudo-inverse operator. Then, an eigendecomposition is performed  on $\mathbf{A}_{batch}$ to estimate the DMD modes and eigenvalues. For low-dimensional systems and short time scales, the performance of the systems are similar. For an environment with dimension $20 \times 20$ and a complex spatiotemporal process simulated $1000$ time steps over $10$ seconds, the dominant eigenvalue is such that $\mathrm{Im}(\lambda_{\text{DMD}}) = 0$ and $\mathrm{Re}(\lambda_{\text{DMD}}) = -1$. The data is simulated and injected with Gaussian random noise $\mathcal{N}(0, \sigma^2)$ for $\sigma^2=0.01$, $0.04$, and $0.1$. The online DMD methods are first initialized with data collected over $400$ time steps and updated every $10$ time steps. For each $\sigma^2$ value, 10 trials are collected and the data is aggregated to analyze their descriptive statistics. Fig. \ref{fig:eig-val-comparison} shows the comparison of the naive batch algorithm, the general adaptation method, and long-term adaptation method. The boxplots represent the median, upper and lower quantiles, and outliers for each online DMD method under low, medium, and high noise regimes. While the online methods do not converge exactly to the eigenvalue, they provide close, fast approximations when compared to the batch method. After the initial DMD model is computed, convergence for all three methods occurs after several updates, around $t = 5s$. For all the three methods, the greater the noise, the greater the spread of the estimates of the eigenvalue across trials. The general method provides closer estimates of the real parts of the eigenvalue, while the long-term method provides closer estimates of the imaginary parts of the eigenvalue. The long-term methods exhibits greater variance in its estimates compared to the general method.

The general adaptation scheme allows for computation of eigenmodes using the low-rank operator $\hat{\mathbf{A}}$, as in Eq. \eqref{eq:projected-A}. This method is computationally efficient in a wide range of scenarios, but does require storing the previous SVD of the data matrix $\mathbf{X}$ and all of the snapshots of the data in order to compute the low-rank operator. There are certain conditions under which this operator is unable to produce accurate estimates of the operator $\hat{\mathbf{A}}$. As presented by \cite{Tu2013OnApplications}, the eigendecomposition of the operator $\hat{\mathbf{A}}$ produces the eigenvalues and eigenvectors of $\mathbf{A}$ only when $\mathbf{Y}$ lies in the span of $\mathbf{X}$. Otherwise, the eigenvalues and eigenvectors of the low-dimensional operator are useless. While the authors suggest this is easily fulfilled in dynamical systems, in reality, this is not a condition that is known a priori and may lead to extremely poor estimates of the dynamical system. For example, if a system is slowly varying between time steps where $\mathbf{Y}$ is simply a time-shifted $\mathbf{X}$, it may very well be that $\mathbf{Y}$ could be written as a linear combination of the matrix $\mathbf{X}$ and therefore $\mathbf{Y}$ would lie in the span of $\mathbf{X}$. In general, for nonlinear systems this should not be common as this would suggest the matrix $\mathbf{Y}$ is not a simple linear combination of the matrix $\mathbf{X}$. However, in practice, some downsampling in the time domain may be necessary to use the algorithm.

The long-term modeling scheme directly computes the $\mathbf{A'}$ operator and allows for the inclusion of a weighting parameter. The weighting parameter, $\gamma$, is advantageous for disregarding the previous data that may not capture the relevant dynamics, as shown in Fig. \ref{fig:online-weighting-comparison}. Allowing for a forgetting factor improves the estimate of the DMD spatial and temporal modes. While this work was proposed by \citep{Zhang2019OnlineSystems}, there does not exist, to the best of our knowledge, a systematic way for picking $\gamma$ to ensure the performance of the system. It has not been shown whether under certain conditions, values of $\gamma$ allow for a convergence to the true DMD operator. Though computing the operator $\mathbf{A'}$ is advantageous in terms of model accuracy, it comes with some caveats. First, this method requires initialization with full rank matrices. This means that in order for the DMD computation to be initialized at least $N$ data points will have to have been collected for the stability of the algorithm. Otherwise, the matrix will be ill-conditioned and the model will be erroneous. For high-dimensional systems, this constraint is hard to fulfill. For example, for a $100 \times 100$ environment, $N$ would equal $10000$, and thus, at least, $10000$ time steps of data would be required for the algorithm. This may not be feasible in all applications. It is for this reason that in Sect. \ref{sec:experimental-validation}, we were unable to test the algorithm experimentally on a high-dimensional system projected onto the water tank where we only had $\sim 100$ time steps of data. Second, there is additional computational time associated with the construction of the matrix and then its eigendecomposition, instead of directly computing its eigenvalues and eigenvectors.

Table \ref{tab:exec-time} demonstrates the execution times of the general and long-term online adaptation methods compared with the batch algorithm. All of the timing calculations were performed on the same synthetic data but with various discretizations in time and dimension. The system is defined as 
\begin{equation*}
    f(x, y, t) = senh(x)senh(y)*1.9i^{-t},
\end{equation*} where $senh(z) = (e^{z} + e^{-z})/2$. The system is a hyperbolic sine with damped oscillation. The data is simulated by using the equation and injecting Gaussian random noise $\mathcal{N}(0, 0.4)$ for each trial. Additionally, the parameters $T$, the initial batch size of the data used to compute the DMD model, and the parameter $\tau$, the batch size of the assimilated data, are varied across environments. Note, the parameter $T$ depends on the size of the dimension for the long-term algorithm, as mentioned earlier. For our implementation of the online adaptation algorithms, we used the eigendecomposition algorithms from the scipy Python library, and modified the DMD functions from the PyDMD \citep{Demo2018PyDMD:Decomposition} library. All computations were performed on a Dell Optiplex 9020 with a 3.40 GHz Intel i7-4770 CPU. 

The batch algorithm and long-term algorithm compute estimates of the operator $\mathbf{A}$ as specified by their respective algorithms. The general algorithm, however, directly computes the eigenvalues and eigenmodes. For the comparison of the online DMD methods with eigendecompositions, the batch and long-term algorithms must perform an eigenvalue decompositions, while the general algorithm merges the eigenvectors into a DMD basis. Thus, the batch algorithm must incrementally perform both the pseudo-inverse operation and eigendecomposition with increasingly larger matrices. Similarly, the long-term algorithm must perform the eigendecomposition on increasingly larger matrices. 

For small environments and short time scales, as in the case of the $10 \times 10$ environments, there is negligible difference between the general and long-term algorithms. Both of them outperform the batch algorithm, as they do not have to perform a pseudoinverse and subsequent matrix decomposition operator during each assimilation step. In this scenario, even though the long-term algorithm requires explicit eigendecomposition of the estimated $\mathbf{A'}$ matrix, it avoids the repeated SVD computations required of the general algorithm. This results in the faster performance of the long-term algorithm compared to the general algorithm for environments with small dimensions and short duration. The computations associated with the eigendecomposition take less time than the various low-dimension matrix operations for environments for small dimensions. For larger environments with longer durations, such as the $20 \times 20$ with $1000$ time steps, the general adaptation method outperforms long-term adaptation method both with and without the SVD computation. The online algorithm is not well-suited for large environments. For smaller environments ($\sim200$ spatial points) being monitored over a long duration, such as the $20 \times 10$ grid with $2000$ time steps, the long-term algorithm is faster than the general algorithm without eigendecomposition. It is worth noting that the long-term algorithm does not require storage of any past snapshots, while the general algorithm does, which in the case of long-term monitoring can become prohibitively large. For the general algorithm used for monitoring of high-dimensional system in \ref{subsec:simulation-barium}, this can be up to $3 MB$, while for the long-term monitoring it is only several hundred kilobytes.

\section{Mixed reality experiments}
\label{sec:experimental-validation}
\begin{subsection}{Simulation of proposed approach on barium cloud data}\label{subsec:simulation-barium}

\begin{figure*}
\centering
\subfloat[ ]{
  \includegraphics[width=53mm]{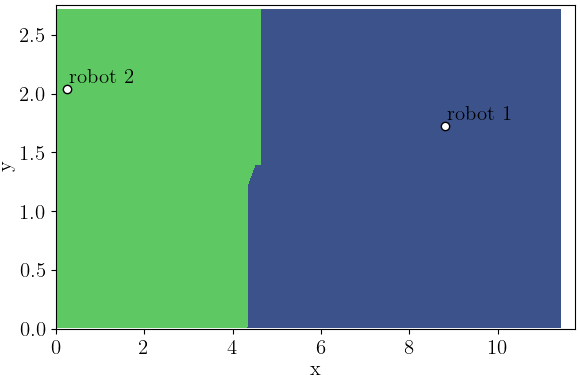}
}
\subfloat[ ]{
  \includegraphics[width=63mm]{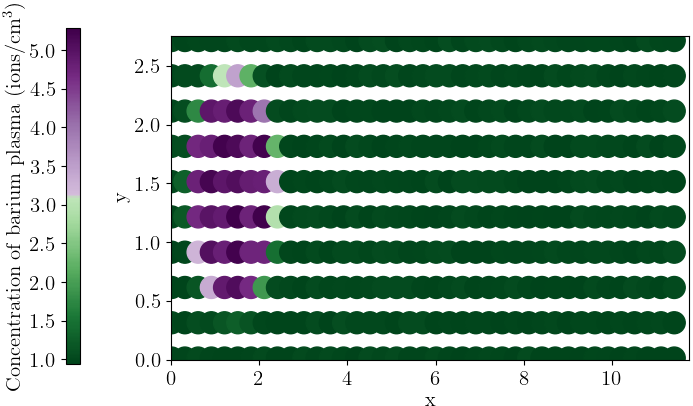}
}
\subfloat[ ]{
  \includegraphics[width=53mm]{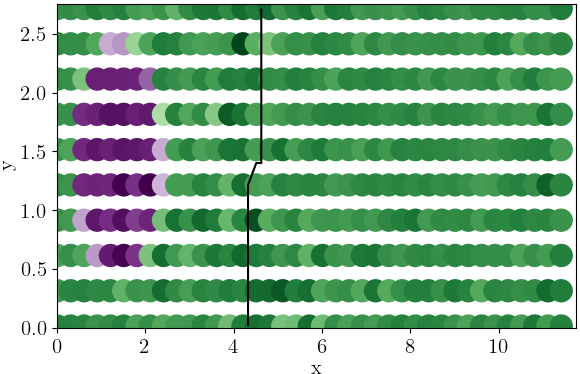}
}
\caption{Depiction of coverage control technique and sensing quality for aerial vehicles. (a) Aerial vehicles are randomly assigned sensing to begin. Locations are assigned to robots for coverage based on their proximity to the robot. (b) The true field can be sensed at the spatial locations shown. (c) The robots each take one sensing region, where the regions are split by the black line. The further away a location within a robot's sensing region is from the robot's position, the greater the noise in the data associated with that position. Thus, quality of the sensor measurements deteriorates further away from the position of the robots, as seen by the differences in colors from robots' sensing measurements and the true field}\label{fig:cvg-ctrl-exp}
\end{figure*}

\begin{figure}[ht!]
    \centering
    \subfloat[ ]{\includegraphics[width=0.46\textwidth]{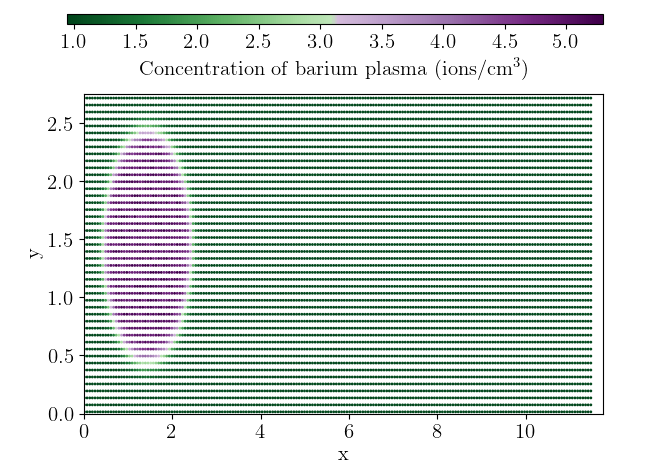}}
    
    \subfloat[ ]{\includegraphics[width=0.46\textwidth]{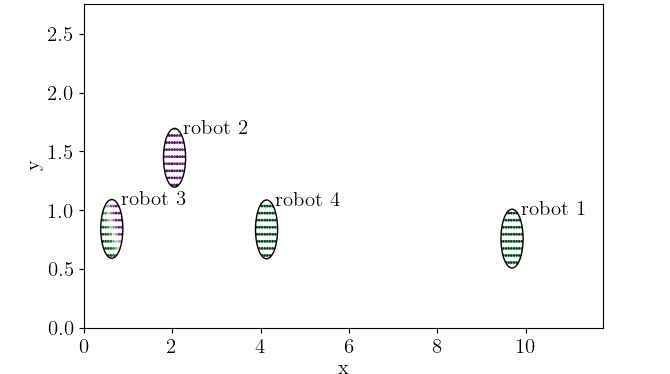}}
    \caption{Depiction of assignment to sensing regions for marine vehicles. (a) The marine vehicles sense the true process at a greater spatial resolution than the aerial vehicles. If the marine vehicles had full coverage of the process, the resulting process would appear as such. (b) However, marine vehicles have limited sensing radius and do not have sensing data available for spatial locations outside of their regions. The circular regions represent the data being sensed by each of the robots, whereas the white space represents regions for which sensing data is unavailable. An estimate of the true process in the white space is constructed using the limited sensing data from robots and a model} \label{fig:sensing-region-exp}
\end{figure}

\begin{table}
\centering
\caption{Comparison of Mean-Squared Averaged Over Entire Spatial Field and Time for Various Simulated Artificial Plasma Clouds and Algorithm Parameters}
\begin{tabular}{|l||c|l|c|l|c||l|}\hline

\multirow{3}{*}{\vspace{-1.1em}{Environment}}&\multicolumn{4}{|c|}{Data Sampling Rate}&\multicolumn{1}{|c||}{}&\multicolumn{1}{|c|}{}\\ \cline{2-5}
&\multicolumn{2}{c|}{Time}&\multicolumn{2}{c|}{Space}&Noise&MSE\\ \cline{2-3}\cline{4-5}
&AV&MV&AV&MV&&\\ \cline{1-7}

\hline \hline \multirow{3}{*}{Cloud 1}
&1&5&12&4&0.15&1.014\\ \cline{2-7} & 1&5&9&3&0.12&0.9885\\ \cline{2-7} & 1&5&6&2&0.12&0.9709 \\ \cline{1-7}

\hline \hline \multirow{3}{*}{Cloud 2}
&1&5&12&4&0.15&0.0968 \\ \cline{2-7} &1&5&6&2&0.12&0.0947 \\ \cline{2-7} & 1&2&9&3&0.12&0.1235\\  \cline{1-7}

\hline \hline \multirow{2}{*}{Cloud 10}
&1&5&12&4&0.15&3.078\\ \cline{2-7} &1&2&9&3&0.12&3.166 \\ \cline{1-7}

\end{tabular}
\label{tab:clouds-comp}
\end{table}

\begin{figure}[ht!]
    \centering
    \subfloat[ ]{\includegraphics[width=0.46\textwidth]{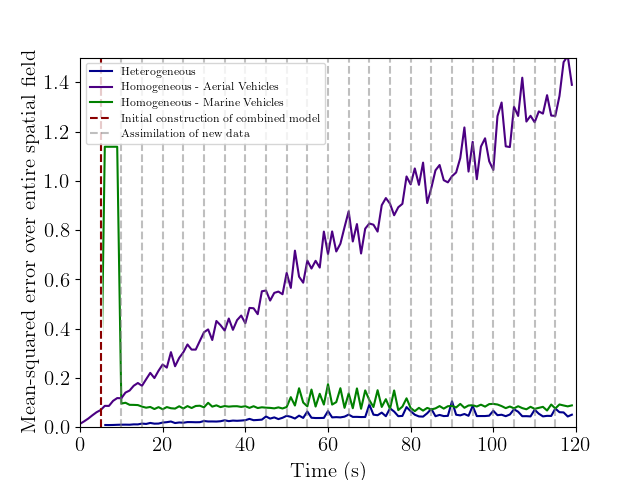}}
    
    \subfloat[ ]{\includegraphics[width=0.46\textwidth]{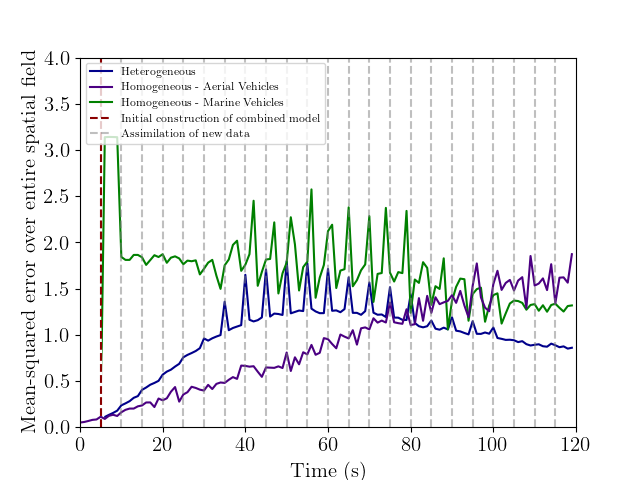}}

    \caption{Error over time comparing models from heterogeneous data and homogeneous data. Mean-squared error over entire spatial grid is calculated over the time series for reconstruction from heterogeneous data, homogeneous data from aerial vehicles, and homogeneous data from marine vehicles. (a) The simulated environment, while nonlinear, is relatively simple compared to the other tested environments. The model with heterogeneous outperforms both of the homogeneous models for the duration of the process. (b) For a more complicated process, the estimates from the aerial vehicle are better to begin. However, as the process continues, the heterogeneous model estimates result in lower error.} \label{fig:mean-error-sim}
\end{figure}

In this section, we show that using our proposed algorithm,the performance of the heterogeneous system is better than that of a homogeneous system and that the optimal sensing locations derived from our proposed algorithm impact the overall quality of the model in complex environments. A simulated environment is created using the density of an artificial plasma cloud in the near-Earth space environment. The density of this artificial plasma cloud was derived in simulation, using the models presented by \cite{Zalesak1987DynamicsClouds}. These processes do not simply convect, but rather can become unstable and undergo complex nonlinear evolution. When used as an environmental processes of interest, the evolution of the density of the artificial plasma cloud allows us to study the ability of the robots to track an unstable, complex nonlinear phenomena. 

We compare various clouds on different algorithm parameters and study their average over time. The clouds are computed using computation fluid dynamics models of artificial plasma clouds that serve as representation for barium clouds in the atmosphere. These models were initialized under various parameters from the equations in \cite{Zalesak1987DynamicsClouds}. To begin, the simulated data is collected and interpolated by the simulated aerial vehicles for a certain predetermined number of time steps. This is used to create an initial model using DMD. The marine robots' positions are randomly initialized. Then, the sensing data from the marine robots is used to estimate the full field and combined into the model. The spatial and temporal characteristics are extracted from the model. Based on this initial information, the robots are sent to new optimal sensing locations. This is done using the methods described in Sections \ref{subsubsec:block-qr} and \ref{subsubsec:coverage} for the marine vehicles, and aerial vehicles, respectively. The data from the aerial vehicle is collected for the next predetermined number of time steps, at which point the data is combined with the high-fidelity sensing data and estimates from the marine vehicle. All of this data is assimilated into the original model using the generalized adaptation techniques. At this point, there is new spatial and temporal characteristics extracted. The process is repeated for the whole time series, at which point the algorithm is terminated.

First, we assume two aerial vehicles have full coverage and five marine vehicles each have a sensing radius of $15$ units. It is assumed aerial vehicles and marine vehicles are collecting sensor measurements using the paradigms shown in Fig. \ref{fig:cvg-ctrl-exp} and Fig. \ref{fig:sensing-region-exp}, respectively. The mean-squared error over the entire spatial grid calculated over the time series as $mse$ is calculated for field estimate $\bm{\hat{x}(t)}$ using $mse = \frac{1}{N} \sum_{i=1}^N \left( \bm{\hat{x}}_i(t) - \bm{{x}}_i(t) \right)^2$. In this case, the model estimates at each location $i$, $\bm{\hat{x}}_i(t)$ is compared against the value of the density from the simulated data, $\bm{{x}}_i(t)$. This is then averaged over all time steps. The results from these simulations are shown in Table \ref{tab:clouds-comp}. The data sampling rates are varied for the aerial vehicle (AV) and marine vehicles (MV). The data sampling rate is an number integer number, $n$, indicates that every $n^{th}$ element was taken from the full simulated data in either the time or space domain. The time sampling rate for the marine vehicle also indicates how after the model is updated. For example for the MV time sampling rate $5$, the model is updated every $5$ time steps, where the marine vehicle only has data at the $5^{th}$ time step. The same noise is injected into both the aerial vehicle and marine vehicle sensor data. 

Using the results from the Cloud $1$ simulation, we see higher noise results in worse estimates, as expected. We also see that greater spatial resolution leads to better results. However, for faster update times, seen in the results from Cloud $2$, where the MV time sampling rate is $2$, we observe worse results from an increase in time sampling. This is due to the fact that early models of the process from few data points are poor and these errors from the early propagate throughout the duration of the process. We also observe very low errors in Cloud $2$, when compares to Clouds $1$ and $10$. This is due to the relative complexity of Cloud $2$ to Clouds $1$ and $10$. Cloud $2$ exhibits much simpler behavior throughout the process.

We also observe the model estimate over the duration of the full process and compare them to models created using just homogeneous data. The estimates of the field are computing by using data from just the aerial vehicles, data from just the marine vehicles, and data combined from both types of vehicles. The mean-squared error is computed over the entire spatial grid for each time step, as shown in Fig. \ref{fig:mean-error-sim}. This corresponds to the simulation in Table \ref{tab:clouds-comp} for row $1$ in the environment using Cloud $2$ in (a) and row $1$ in the environment using Cloud $2$ in (b). The estimates using the data from the aerial vehicles are computed using standard interpolating techniques. The estimates using the data from the marine vehicles are computed by constructing a DMD model from the model, as in Eqs. \eqref{eq:A-decomp} and \eqref{eq:dmd-modes}, and using prediction techniques for times when sensing data is unavailable, as in Eq. \eqref{eq:dmd-reconstructed}. The estimates using data combined from both types of vehicles are computed as described above. Note, that for the DMD model computed using homogeneous marine vehicle data, we assume the homogeneous data is collected from the optimal sensing locations generated by our model. Using random locations results in extremely high errors. The homogeneous data from the aerial vehicles is noisy and collected at a much lower spatial resolution then the true process. As the process becomes more complex, the inclusion of multiple types of data allows the proposed approach to outperform either of the other estimations. This is especially evident for Fig. \ref{fig:mean-error-sim} (b), where towards the end of the process the optimal sensing locations are critical in improving the model performance for the model using heterogeneous data.

\end{subsection}

\begin{subsection}{Experimental validation in water tank test bed}
In the following, we present results to show our method successfully uses a heterogeneous team of robots to model and infer the properties of a complex spatiotemporal phenomena. We evaluate this claim for the general adaptation strategy presented above and also present results on a test bed of real robot to validate the use of the development of these algorithms in physical environments.

\begin{figure*}[ht!]
    \centering
    \subfloat[ ]{\includegraphics[width=0.45\textwidth]{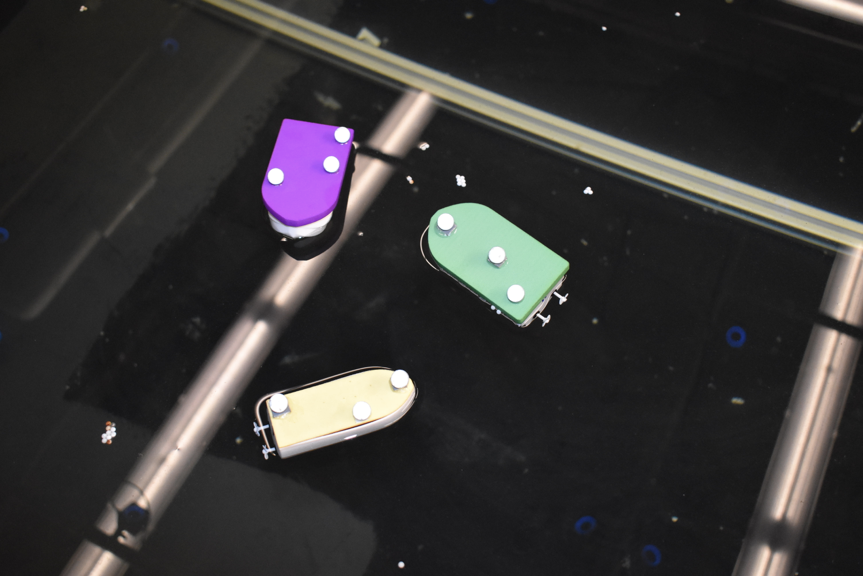}}
~
    \subfloat[ ]{\includegraphics[width=0.45\textwidth]{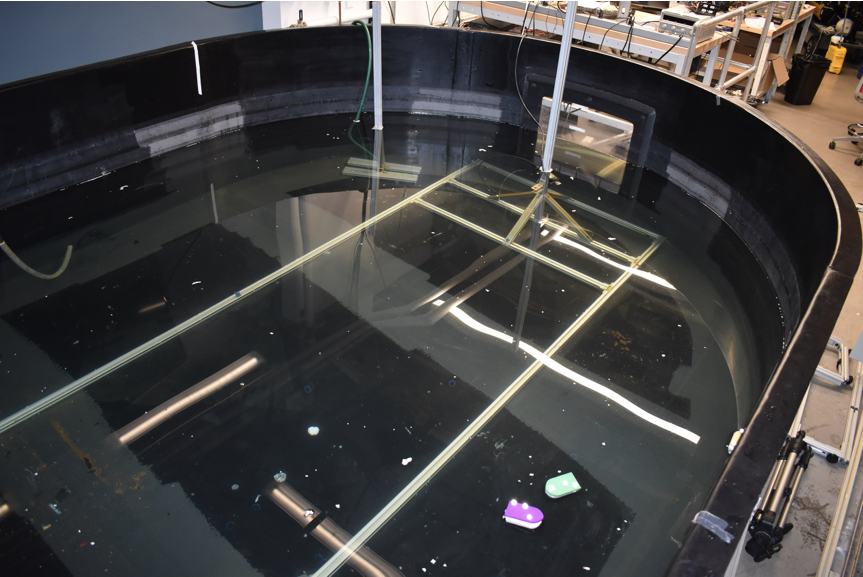}}

    \caption{Experimental setup used for testing algorithms. (a) Micro-autonomous surface vehicles collecting high-fidelity sensor measurements. (b) Indoor $4.5m \times 3.0m \times 1.2m$ water tank equipped with motion capture system for multi-robot experiments}\label{fig:exp-setup}
\end{figure*}

To model and predict properties of a complex, nonlinear process, we use a fleet of three real micro-autonomous surface vehicles (mASVs), one simulated mASV, and two simulated aerial vehicles. The mASVs are differential drive and include a micro-controller board, an XBee radio module, and an inertial measurement unit. The XBee radio modules are used to communicate the information collected onboard. The testbed tank is $4.5m \times 3.0m \times 1.2m$ and is equipped within a motion capture system used for localization. The mASVS are all localized using this motion capture system. This experimental setup is shown in Fig. \ref{fig:exp-setup}. 

We ran experiments on $100$ seconds of data to evaluate our general online adaptation strategy in Sect. \ref{subsec:online-adaptation} and using the environmental setup described in Sect. \ref{subsec:simulation-barium}. The process we are tracking is the density of an artificial plasma cloud in the near-Earth space environment, derived in the same way as the simulated environment described in Sect. \ref{subsec:simulation-barium} and simulated using computational fluid dynamics models based on the equations in \cite{Zalesak1987DynamicsClouds}. This process is projected onto the tank as an image, where the pixel values of the image are mapped to concentration values of the density of the artificial plasma cloud for the aerial vehicle. We use simulated sensing data as noisy sparse measurements from the concentration data for marine vehicles.  In simulation, travel time is not explicitly accounted for in the determination of optimal sensing locations. Instead, it is assumed that the robots will be able to travel to these locations before the next adaptation time. This is validated in these experiments, where robots are in fact able to determine optimal sensing locations, assign robots in the team to locations based on minimizing cumulative time, travel to optimal sensing locations, and collect sensor measurements all before the next adaptation time.

We evaluate the full framework with the generalized adaptation framework. The environment is originally defined over a $384 \times 768$ workspace. In this environment, we have four marine robots and two aerial vehicles. The workspace is down-sampled to a $10 \times 20$ grid for the low-fidelity data. This low-fidelity data is simulated as on-board sensing data for the aerial vehicles. Aerial vehicles are assumed to have full views of their sensing region as determined by a coverage control method, where the quality of their sensing data diminishes at points further away from their current sensing location, as in Fig. \ref{fig:cvg-ctrl-exp}. This is simulated by calculating the desired position of the aerial vehicles from the coverage control technique and adding Gaussian random noise to the true sensing data as a function of the distance from the desired position. This data is then interpolated to the dimensions of the high-fidelity data collected by the marine robots which is on a $48 \times 96$ grid. The marine robots have a sensing radius of $10$, meaning that within the $48 \times 96$ grid, a marine robot can collect sensing information over all of the grid points laying in a circle with radius $10$ centered at its position, as in Fig. \ref{fig:sensing-region-exp}. The high-fidelity sensing data from the marine vehicles is combined into a single vector $\mathbf{x}_L$ and used to extrapolate the data from the $48 \times 96$ grid using Eq. \eqref{eq:gappy-pod}, based off of Gappy POD methods \citep{Everson1995Karhunen--LoeveData, Manohar2018Data-DrivenPatterns}. While the aerial vehicles are able to collect sensor measurements across the full time series, marine vehicles are only able to collect data periodically at regular intervals.

\begin{figure*}
\centering
\subfloat[ ]{
  \includegraphics[width=0.45\textwidth]{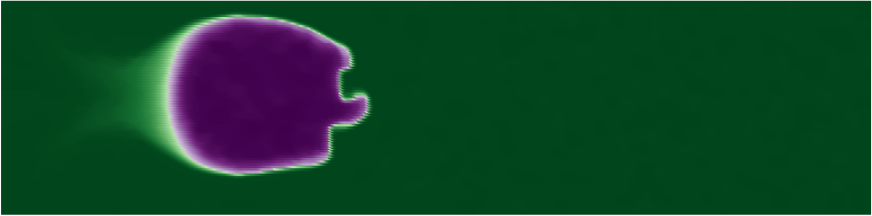}
}
\subfloat[ ]{
  \includegraphics[width=0.45\textwidth]{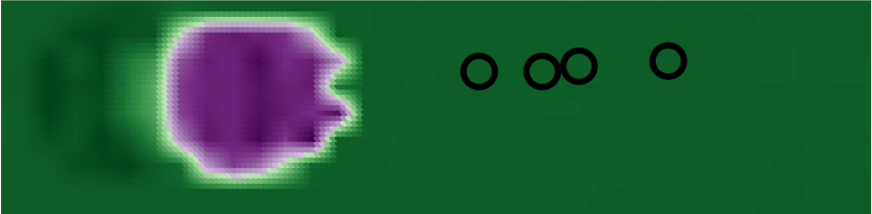}
}
\\
\subfloat[ ]{
  \includegraphics[width=0.45\textwidth]{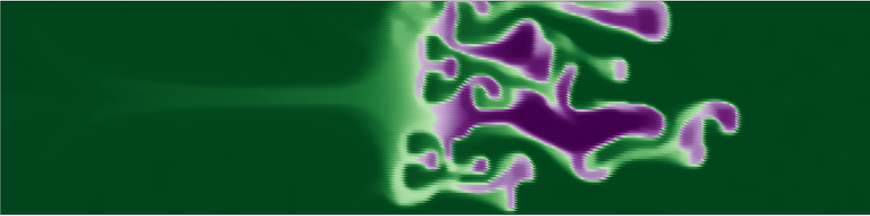}
}
\subfloat[ ]{
  \includegraphics[width=0.45\textwidth]{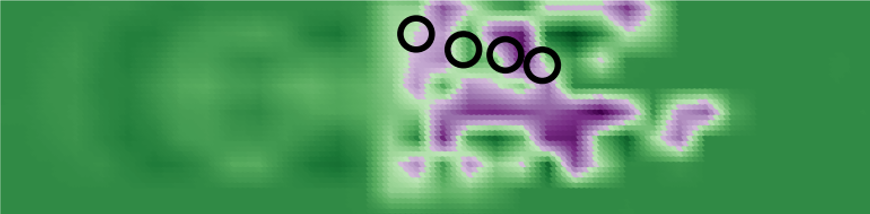}
}
\caption{Depiction of field reconstruction using adapted DMD models and sensor measurements from robots. (a) True field at full spatial resolution as obtained by artificial plasma cloud simulation at time step $40$. (b) Reconstructed of field using heterogeneous data at time step $40$, with marine robot locations depicted by black circles. (c) True field at full spatial resolution as obtained by artificial plasma cloud simulation at time step $40$. (d) Reconstruction of field using heterogeneous data at time step $40$, with updated marine robot locations}
    \label{fig:adaptation-over-time}
\end{figure*}

An example of the robots tracking an artificial plasma cloud is shown in Fig. \ref{fig:adaptation-over-time}. In the beginning of the experiment, as in Fig. \ref{fig:adaptation-over-time} (a), the robots are able to create an reconstruction of the field, shown by Fig. (b). It should be noted, the robots are taking sensor measurements in the areas of the process such that they are not collecting varied information, as seen in Fig. (b). However, as the experiment continues and the robots are able to collect more data, the model is improved, such that the optimal sensing locations correspond to the areas of interesting dynamics in the process, as evident in Fig. \ref{fig:adaptation-over-time} (c) and (d). It is evident that the reconstruction is able to capture the complex, nonlinear dynamics of the process, shown in Fig. \ref{fig:adaptation-over-time} (d). More specifically, we see robots collected sensor measurements at or near the spatial locations that correspond to the complex phenomena in the process. The robots in the mixed reality experiment are able to successfully track a complex, nonlinear environmental process using their heterogeneous multiscale, multiresolution data.
\end{subsection}

\section{Conclusion}
\label{sec:conclusion-future}
In this work, we contribute to the understanding of using heterogeneous robot systems for fusion of information to the end of modeling complex and multiscale nonlinear phenomena. We consider the problem of leveraging multiresolution sensor information from a heterogeneous robot team to model and predict the evolution of a spatiotemporal process. Using these models, we infer the optimal sensing locations using computationally efficient methods. We show that the collection of this information and adaptation of the models with measurements from disparate sources results in accurate reconstructions of the estimated field. The frameworks presented allows for a decoupling of the temporal and spatial modes apparent in the data. This decoupling is then used within a task allocation framework for the various types of robots. Instead of relying on the standard task-trait allocation approaches typically used in heterogeneous robotic framework, this approach leverages the unique strengths of the robots to jointly complete a task. Additionally, the framework proposes two distinct ways for assimilating online data. In the case of data assimilation for long-term modeling, the forgetting factor is useful in scenarios where the environmental process is quickly time varying. However, the precise nature of the forgetting factor, specifically what the optimal value is for a specific process, is not well understood and requires greater investigation \citep{Zhang2019OnlineSystems}. The reliance of this framework on DMD and its inherent capability for future state-estimation, as in Eq. \eqref{eq:dmd-reconstructed}, may allow for the use of other optimal sensing techniques that account for not only the current dynamics of the system but also the future dynamics of the system. Much of the literature in the intersection of optimal sensing and robotics rely on the use of GPs. However, in these scenarios, the spatial dimension of the system, $N$ is extremely high ($\sim 10,000$ points). For modeling time series data, GPs require knowledge of the kernel and the complexity of the algorithm would be $\mathcal{O}^{N^3}$, which is prohibitively large in this scenario. This is contrast to standard DMD methods that have few assumptions and a computational complexity of $\mathcal{O}^{NTk}$ \citep{Erichson2019CompressedModeling}. This work presents and experimentally validates a framework to combine heterogeneous sensing and mobility capabilities to learn, optimally sense with respect to, and adapt a model.


%
%

\bibliographystyle{spbasic}      
\bibliography{references}   

\end{document}